\documentclass[journal]{IEEEtran}
\usepackage[american]{babel}
\usepackage{microtype}
\usepackage[pdftex]{graphicx}
\usepackage{subfigure}
\usepackage{amsthm}

\usepackage{amsmath, amsfonts}

\usepackage{algorithm}
\usepackage{algorithmic}
\usepackage{enumerate}
\usepackage{flushend}
\usepackage{booktabs}
\usepackage{longtable}
\usepackage{setspace}
\usepackage{verbatim}
\usepackage{url}
\usepackage{cite}
\usepackage{color}

\usepackage{ulem}
\usepackage{float}
\usepackage{makecell}
\usepackage{multirow}

\begin{document}
\title{SDA-$x$Net: Selective Depth Attention Networks for Adaptive Multi-scale Feature Representation}

\author{
\author{Qingbei~Guo\IEEEauthorrefmark{1},
        Xiao-Jun~Wu\IEEEauthorrefmark{2},
        Zhiquan~Feng\IEEEauthorrefmark{1},
        Tianyang~Xu\IEEEauthorrefmark{2} and
        Cong~Hu\IEEEauthorrefmark{2}
\thanks{Q. Guo and Z. Feng are with with Shandong Provincial Key Laboratory of Network based Intelligent Computing, University of Jinan, Jinan 250022, China. (e-mail: \{ise\_guoqb; ise\_fengzq\}@ujn.edu.cn)}
\thanks{X.-J. Wu, T. Xu and C. Hu are with the School of Artificial Intelligence and Computer Science, Jiangnan University, Wuxi, P.R. China. (e-mail: \{xiaojun\_wu\_jnu; tianyang\_xu\}@163.com; conghu@jiangnan.edu.cn)}
}
}





\markboth{IEEE Transactions on Image Processing (Submitted)}
{Shell \MakeLowercase{\textit{et al.}}: Bare Demo of IEEEtran.cls for IEEE Journals}

\maketitle


\begin{abstract}
Existing multi-scale solutions lead to a risk of just increasing the receptive field sizes while neglecting small receptive fields.
Thus, it is a challenging problem to effectively construct adaptive neural networks for recognizing various spatial-scale objects.
To tackle this issue, we first introduce a new attention dimension, i.e., depth, in addition to existing attention dimensions such as channel, spatial, and branch, and present a novel selective depth attention network to symmetrically handle multi-scale objects in various vision tasks.
Specifically, the blocks within each stage of a given neural network, i.e., ResNet, output hierarchical feature maps sharing the same resolution but with different receptive field sizes.
Based on this structural property, we design a stage-wise building module, namely SDA, which includes a trunk branch and a SE-like attention branch.
The block outputs of the trunk branch are fused to globally guide their depth attention allocation through the attention branch.
According to the proposed attention mechanism, we can dynamically select different depth features, which contributes to adaptively adjusting the receptive field sizes for the variable-sized input objects.
In this way, the cross-block information interaction leads to a long-range dependency along the depth direction.
Compared with other multi-scale approaches, our SDA method combines multiple receptive fields from previous blocks into the stage output, thus offering a wider and richer range of effective receptive fields.
Moreover, our method can be served as a pluggable module to other multi-scale networks as well as attention networks, coined as SDA-$x$Net.
Their combination further extends the range of the effective receptive fields towards small receptive fields, enabling interpretable neural networks.
Extensive experiments demonstrate that the proposed SDA method achieves state-of-the-art (SOTA) performance, outperforming other multi-scale and attention counterparts on numerous computer vision tasks, e.g., image classification, object detection, and instance segmentation.
Our source code is available at \url{https://github.com/QingbeiGuo/SDA-xNet.git}.
\end{abstract}


\IEEEpeerreviewmaketitle


\section{Introduction}\label{sec:Introduction}

\begin{figure}[t]
  \centering
  \includegraphics[trim=0mm 0mm 0mm 0mm, width=3.0in]{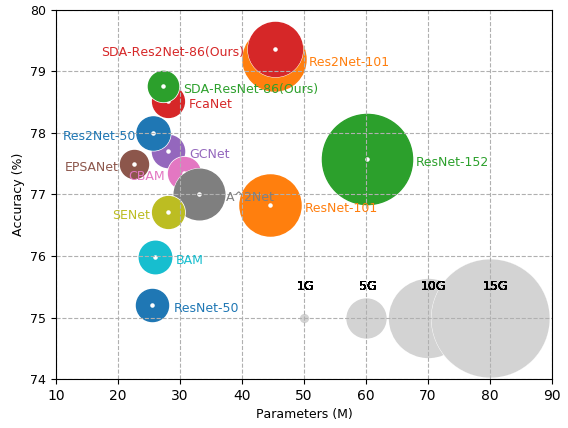}
  \caption{Image classification performance comparison of different multi-scale and attention methods with ResNet-50 backbone on ImageNet-1k in terms of Top-1 accuracy, computational cost, and model size. The scale of the circles indicates  GFLOPs. The proposed SDA-ResNet-86 outperforms a much larger ResNet-152 architecture on the ImageNet-1k dataset with over 2$\times$ less the number of parameters and close to 3$\times$ lesser FLOPs.}
  \label{fig:1}
\end{figure}

The involvement of multi-scale feature representations is of great practical significance for various computer vision (CV) tasks, including image classification~\cite{KrizhevskySH12}, object detection~\cite{RenHGS15}, semantic segmentation~\cite{ChenPKMY17}, and instance segmentation~\cite{HeGDG17}, etc., due to diverse sized target objects.
Therefore, multi-scale feature representations have been widely used in the design of convolutional neural networks (CNN) such as Res2Net~\cite{GaoCZZYT19}, PyConv~\cite{DutaLZS20} and EPSANet~\cite{ZhangZLZM21}.
Among those, the multi-kernel convolution is one of the most commonly used methods, which exploits different kernel sizes or multiple kernel cascades to extract features with different scales for the aggregation of rich multi-scale information.

Although such multi-kernel methods can perceive different scales, they tend to rapidly increase the receptive field sizes, which is a disadvantage when performing recognition for small target objects~\cite{LuoLUZ16,LiCWZ19}.
For instance, the recently proposed EPSANet~\cite{ZhangZLZM21} is a variant neural network of ResNet~\cite{HeZRS16}, which replaces the 3$\times$3 convolution with the Pyramid Split Attention (PSA) module to improve the ability of the multi-scale feature representation.
The PSA module is composed of several groups of convolutions with different kernel sizes, e.g., 3, 5, 7, and 9.
Thus, the network output combines multiple components of different size receptive fields.
The 3$\times$3 convolution contributes to the smallest part of receptive fields, which is equivalent to ResNet in terms of the receptive field settings.
Other convolutions with larger kernels increase the receptive field sizes.
In other words, EPSANet achieves a wide range of effective receptive fields, starting from the smallest part equivalent to ResNet.
However, such large receptive fields prefer to recognize large target objects, while neglecting small target objects.
Therefore, how to design an adaptive multi-scale neural network is the key to extracting a wide range of receptive fields from small to large.

In this work, our goal is to design a novel neural network as an adaptive multi-scale feature selector for the prediction of scale-varying objects.
To this end, we first introduce a new attention dimension, i.e., depth, in addition to existing attention dimensions such as channel, spatial, and branch.
Based on this, we present a novel depth attention network that adaptively fuses the multi-scale features from different depth blocks for various scales of objects.
Concretely, in each stage of neural networks, we build a depth attention module, which consists of two branches, i.e., a trunk branch and an attention branch.
We use the unit of ResNet as the trunk branch, which is responsible for producing the hierarchical features from the blocks of each stage.
The feature hierarchy consists of a series of feature maps with the same number of channels and the same resolution but different receptive field sizes, which is helpful to align for the feature fusion in the channel and spatial dimensions.
Similar to SENet~\cite{HuSS18}, the attention branch is implemented as a lightweight network module, acting as a feature selector for the hierarchical features of the trunk branch.
Global embedded information is obtained by a global average pooling (GAP), followed by two simple Fully-Connected layers (FC), with a softmax operation being applied to obtain the attention weights of the corresponding hierarchical features.
Based on the weighted multiplication and element-wise summation, we get a rich multi-scale feature representation that corresponds to a wide range of receptive fields.
Small receptive field components from previous blocks are included and delivered across stages.
In this way, a cross-block information interaction is built by adaptively adjusting the receptive field sizes of different blocks, and the long-range dependency is captured along the depth direction.
The proposed SDA-Net is constructed by stacking multiple SDA modules, similar to the residual blocks in the ResNet-like manner.
Moreover, our SDA method is orthogonal to other multi-scale and attention methods, thus SDA-Net can be integrated with their neural networks, improving the model performance for image-related vision tasks with a slight computation burden.
Such a combination merges their receptive field ranges, thus increasing the range of effective receptive fields in the direction of small ones.
A wider range of effective receptive fields can significantly boost the performance, enhancing the interpretability of neural networks, which is verified by our experiments.
Extensive experiments on numerous computer vision tasks including image classification, object detection, and instance segmentation, have been conducted in this work.
The results demonstrate that our SDA-Net achieves superior performance to the state-of-the-art multi-scale and attention networks under similar computation efficiency.
Compared to other attention and multi-scale networks, as shown in Fig.~\ref{fig:1}, the proposed SDA-Net provides the best trade-off between accuracy and efficiency.

The proposed SDA method reflects the following advantages:

\begin{itemize}
\item We first analyze a new attention dimension, i.e., depth, in addition to other existing attention dimensions such as channel, spatial, and branch, thus enabling us to capture the long-range dependency along the depth direction.
\item Based on this, we construct a novel SDA-Net adaptively adjusting the receptive field sizes for rich multi-scale feature representations.
\item The proposed SDA method is orthogonal to other multi-scale and attention methods, which extends the range of effective receptive fields towards small receptive fields for better performance with slight computation increase, providing a new scope of interpretation of neural networks.
\item Extensive experiments confirm that our SDA-Net outperforms the state-of-the-art multi-scale and attention networks on numerous computer vision tasks, i.e., image classification, object detection, and instance segmentation, under a similar model complexity.
\end{itemize}

The rest of this paper is organized as follows:
we first introduce the related work in Section~\ref{sec:RelatedWork}.
We then present the SDA method in Section~\ref{sec:Methodology}.
Through extensive experiments, we demonstrate that the proposed SDA-Net delivers superior performance on numerous computer vision tasks in Section~\ref{sec:Experiments}.
In Section~\ref{sec:Ablation_Study}, we conduct an ablation study to investigate the effect of different factors on our SDA-Net.
Subsequently, we analyze the computational cost, the range of effective receptive fields, the multi-scale representation adaptability, and class activation mapping (CAM) in Section~\ref{sec:Analysis_and_Interpretation}.
Finally, we draw the paper to a conclusion in Section~\ref{sec:Conclusion}.


\section{Related Work}\label{sec:RelatedWork}

\noindent
\textbf{Multi-scale feature representations.}
It has been widely studied that multi-scale feature representations are essential for various computer vision tasks.
TridentNet~\cite{LiCWZ19} utilized the dilated convolution with different dilation rates to generate multiple scale-specific feature maps to solve the scale variation problem of input objects in object detection while exploring the relationship between the receptive field and the scale variation.
A feature pyramid network (FPN) was built by exploiting the inherent multi-scale feature hierarchy of CNNs for object detection~\cite{LinDGHHB17}.
A multi-scale attention model is constructed with multi-scale input images for the performance improvement of semantic segmentation~\cite{ChenYWXY16}.
Some works designed multiple branch networks to aggregate multi-scale information in different scale branches, achieving strong representational power for image classification~\cite{LiKCZ19,ChenFMSF19}.

Multi-kernel convolution is one of the commonly used methods for multi-scale feature representation.
The early InceptionNets~\cite{SzegedyLJSRAEVR15,SzegedyVISW16,SzegedyIVA17} used the inception module with different kernel sizes to extract multiple scale size feature maps for improving the representation ability of models in image classification.
PyConv~\cite{DutaLZS20} and MixNet~\cite{TanL19} exploited the lightweight group convolution with multiple kernel sizes for better model accuracy and efficiency.
Res2Net~\cite{GaoCZZYT19} designed a novel building block with hierarchical residual-like patterns to capture in-layer multi-scale features, increasing the range of receptive fields.
SKNet~\cite{LiWHY19} introduced a novel selective kernel convolution to adaptively adjust different kernel selections in a soft-attention manner.
Similarly, EPSANet~\cite{ZhangZLZM21} used different kernel sizes to extract rich spatial features to achieve strong multi-scale representation ability.

In this work, we exploit the inherent multi-scale representation of CNNs to extract the hierarchical features instead of multi-kernel convolutions, and the small receptive fields of low layers are concerned for a wide range of multi-scale representations.
Furthermore, the proposed SDA-Net differs from FPN, which also uses the multi-scale feature hierarchy of CNNs, because the effective reception field is adaptively captured in the in-stage pattern by using a novel depth attention mechanism.

\noindent
\textbf{Attention mechanisms.}
The attention mechanisms enable neural networks to strengthen the allocation of the most relevant features of input objects while weakening the less useful ones.
Many attention methods have been proposed to explore the attention space.
According to the difference in attention dimensions, these methods can be categorized into three distinct groups: channel attention, spatial attention, and branch attention for image-related vision tasks.
Note that the temporal-wise attention is beyond the scope of this paper.

SENet~\cite{HuSS18} proposed a channel attention method, which adaptively adjusts the channel-wise features by modeling the cross-channel dependency with the squeeze and excitation (SE) operations.
ECANet~\cite{WangWZLZH20} introduced a local cross-channel attention method to reduce the attention network complexity.
GSoP~\cite{GaoXWL19} introduced higher-order global information embedding methods, i.e., global second-order pooling, to model the pair-wise channel correlations of previous image information.
FcaNet~\cite{QinZWL20} proposed another global information embedding method instead of global average pooling in the attention module, which exploits the discrete cosine transform (DCT) to incorporate different frequency components, guiding the selection of channel-wise features.

To overcome the limitation of local convolutions, existing attempts propose to capture the long-range feature interactions at the spatial-wise attention stage.
AANet~\cite{BelloZVSL19} proposed attention augmented convolutional network by combining both convolution and self-attention.
GCNet~\cite{CaoXLWH19} modeled the global context by unifying both the simplified non-local (NL) block~\cite{WangGGH18} and the SE block~\cite{HuSS18}.
A$^2$Net~\cite{ChenKLYF18} proposed a double attention block to capture the long-range spatial feature dependencies via the global information gathering and distribution functions.

The branch attention networks, such as SKNet~\cite{LiWHY19} and EPSANet~\cite{ZhangZLZM21}, used a dynamic branch selection mechanism to adaptively adjust the sizes of their receptive fields according to the variable-sized input objects.
AFFNet~\cite{DaiGOWB21} proposed a multi-scale attention module, which fuses the features from different layers or branches by aggregating local and global contexts.
Similarly, ResNeSt~\cite{ZhangWZZ22} presented a multi-path architecture to capture cross-channel feature interactions by leveraging the split-attention mechanism.

Besides, the combination of two or more attention mechanisms has been also discussed recently, achieving further improvement compared to the use of a single attention mechanism.
For instance, SANet~\cite{ZhangY21} adopted an efficient shuffle attention module to combine channel attention and spatial attention effectively for additional performance gains.
Similar works such as CBAM~\cite{WooPLS18} and BAM~\cite{ParkWLS18} also proposed to combine channel attention and spatial attention for significant performance improvements.

Recently, the transformer architecture has obtained wide attentions in many CV tasks, such as ViT~\cite{DosovitskiyBKWZUDMH20}, DETR~\cite{CarionMSUKZ20} and Deformable DETR~\cite{ZhuSLLWD20}.
But the self-attention mechanism they used is out of the scope of this paper.

In this work, we focus on a novel depth attention method to explore the attention space in the depth dimension, aiming at promoting further research on the attention mechanisms.


\section{Methodology}\label{sec:Methodology}

\subsection{Selective Depth Attention Module}\label{sec:Problem_Definition}

\begin{figure}[t]
  \centering
  \includegraphics[trim=0mm 5mm 0mm 5mm, width=3.50in]{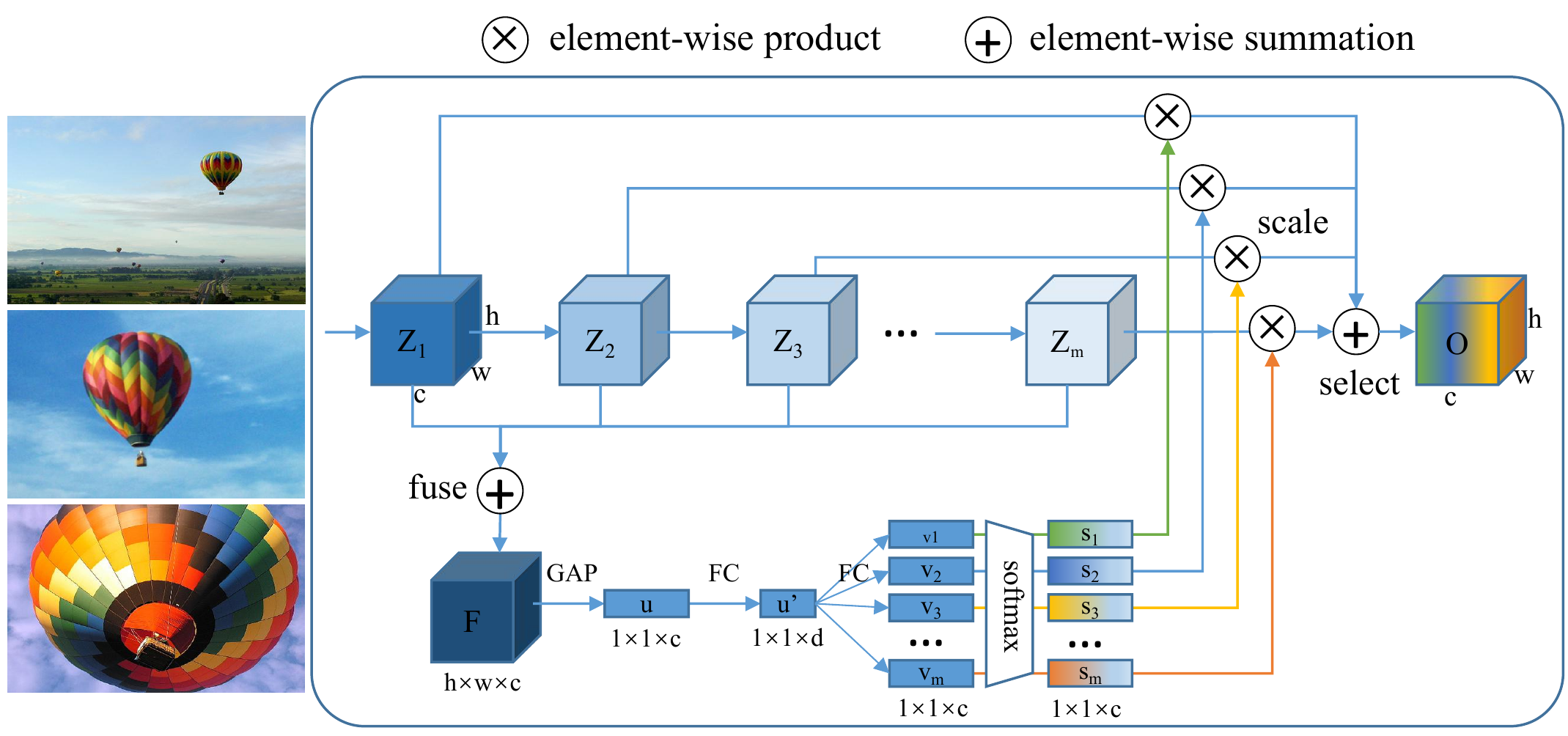}
  \caption{Selective Depth Attention Module.}
  \label{fig:2}
\end{figure}

SDA module consists of two branches: trunk branch and attention branch, as shown in Fig.~\ref{fig:2}.
The trunk branch is driven to learn features based on the typical network architectures.
This approach is independent of the backbone architecture of CNNs, and we use ResNet~\cite{HeZRS16}, SENet~\cite{HuSS18}, and Res2Net~\cite{GaoCZZYT19} as the building unit of our SDA-Net in this paper.
The bottleneck blocks of these three networks are divided into multiple stages according to the difference in output feature resolutions.
The block sequence of each stage serves as the trunk branch of our SDA module.
These blocks output the same number of feature maps with the same resolution but have different receptive field sizes.
It allows us easily merge the multi-scale features from the intermediate blocks, without the problem of spatial alignment.

The attention branch is implemented as a SE-like lightweight network structure.
This branch gathers the feature information from the trunk branch, adaptively weighs the features of the intermediate blocks in the depth dimension, and then passes the resulting multi-scale features into the next stage.
Unlike EPSANet which extracts the multi-scale features using the branch attention mechanism with different size kernels, we exploit the intrinsic feature pyramid representation of CNNs and explore an alternative attention mechanism, i.e., depth attention, for strong multi-scale feature representation capacity.
Thus, the attention branch is used as a feature selector of the trunk branch, which decides the attendance of involved blocks in the feature learning, based on the requirement of feature scales.
It is worth noting that we creatively explore a new depth dimension for the attention space, differently from other existing attention dimensions such as channel, spatial, and branch.

Specifically, for a given SDA module, its trunk branch extracts a sequence of feature maps $\mathbf{Z}=\left[\mathbf{Z}_1, \mathbf{Z}_2, ..., \mathbf{Z}_m\right]$ from the intermediate blocks.
Here, $m$ is the number of blocks, and $\mathbf{Z}_i\in \mathcal{R}^{h\times w\times c}$ denotes the output of the $i$-th block, where $h$, $w$ and $c$ are height, width and the number of channels, respectively.
Thanks to the same number of channels and the same resolution but different receptive field sizes, $\mathbf{Z}$ is essentially a feature hierarchy, which has gradual semantics from low to high levels.
Our goal is to adaptively weight the feature hierarchy via the attention branch to achieve the multi-scale representation for better predictions.

To achieve this goal, we first merge the hierarchical features by using an element-wise summation, as follows,

\begin{equation}
\label{eqn:1}
\begin{aligned}
   \mathbf{F} = \sum_{i=1}^m \mathbf{Z}_i,
\end{aligned}
\end{equation}
\noindent
where $\mathbf{F}\in \mathcal{R}^{h\times w\times c}$.

Then, we aggregate the spatial information of feature maps $\mathbf{F}$ by using GAP to generate the global spatial context descriptor $\mathbf{u}\in \mathcal{R}^{1\times 1\times c}$.
In other words, $\mathbf{u}$ is generated by shrinking $\mathbf{F}$ through its spatial dimensions $h \times w$.
Specifically, the $k$-th element of the descriptor $\mathbf{u}$ is calculated by

\begin{equation}
\label{eqn:2}
\begin{aligned}
   \mathbf{u}_k = \mathbf{GAP}(\mathbf{F}_{:,:,k}) = \frac{1}{h\times w}\sum_{i=1}^{h}\sum_{j=1}^{w}\mathbf{F}_{i,j,k}.
\end{aligned}
\end{equation}

Subsequently, $\mathbf{u}$ is fed into two 1$\times$1 convolutions, as follows,

\begin{equation}
\label{eqn:3}
\begin{aligned}
   \mathbf{v} = \mathbf{W}_2(\gamma(\beta(\mathbf{W}_1\mathbf{u}))),
\end{aligned}
\end{equation}
\noindent
where $\beta$ and $\gamma$ are the functions of BN~\cite{IoffeS15} and ReLU~\cite{GlorotBB11}, respectively.
$\mathbf{W}_1\in \mathcal{R}^{d\times 1\times 1\times c}$ is a learned linear transformation that maps $\mathbf{u}$ to a low-dimensional space for better efficiency, and $\mathbf{W}_2\in \mathcal{R}^{m\cdot c\times 1\times 1\times d}$ is used to resume the channel dimension. Here, $d = max(c/r, L)$, where $r$ is the reduction ratio and $L$ is the threshold value.

Further, we align $\mathbf{v}$ along the depth dimension to obtain $\mathbf{v}^\top$ by a reshaping operation, and employ the softly weighted mechanism with a softmax activation, as follows,

\begin{equation}
\label{eqn:4}
\begin{aligned}
   \mathbf{s}^\top = \delta(\mathbf{v}^\top) = \frac{exp(\mathbf{v}^\top_i)}{\sum_{i=1}^m exp(\mathbf{v}^\top_i)},
\end{aligned}
\end{equation}
\noindent
where $\delta$ denotes the softmax function, and $\mathbf{s}^\top$ is the set of attention vectors.

Finally, at the end of each SDA module, we adaptively fuse different scales of semantic information by softly weighting the cross-block features according to the scale of input objects. Thus, the output of the SDA module can be formulated as:

\begin{equation}
\label{eqn:5}
\begin{aligned}
   \mathbf{O} = \gamma(\sum_{i=1}^m \mathbf{s}_i \mathbf{Z}_i),
\end{aligned}
\end{equation}
\noindent
where $\mathbf{s}^\top$ is reshaped to return to the alignment $\mathbf{s}$ along the channel dimension, and $\mathbf{O}$ denotes the final output of the proposed SDA module.
Note that the attention model captures the channel-wise relationships across blocks, and decides how much attention to pay to the features at different depths, thus achieving the powerful multi-scale feature representation ability.

\subsection{Neural Network Architecture}\label{sec:Channel-wise_Bidirectional_Importance_Estimation_Criterion}

\begin{table*}[!t]
  \scriptsize
  \caption{Architectures for ImageNet classification. The four columns refer to ResNet-50, SDA-ResNet-86, SDA-SENet-86 and SDA-Res2Net-86, respectively.}
  \label{tab:1}
  \centering
  \setlength\tabcolsep{3pt}
  {
  \begin{tabular}{c|c|c|c|c|c}
    \specialrule{0.10em}{1.0pt}{1.0pt}
    stage  & output          & ResNet-50 & SDA-ResNet-86 & SDA-SENet-86 & SDA-Res2Net-86\\
    \specialrule{0.10em}{1.0pt}{1.0pt}
    conv   & 112$\times$112  & \multicolumn{4}{c}{7$\times$7, 64, stride 2} \\
    \specialrule{0.10em}{1.0pt}{1.0pt}
    pooling& 56$\times$56    & \multicolumn{4}{c}{3$\times$3 max pool, stride 2} \\
    \specialrule{0.10em}{1.0pt}{1.0pt}
    stage1 & 56$\times$56    & $\begin{array}{ll} \left[ \begin{array}{llll} 1\times 1, 64 \\ 3\times 3, 64 \\ 1\times 1, 256 \end{array} \right] \times 3 \end{array}$
                             & $\begin{array}{ll} \left[ \begin{array}{llll} 1\times 1, 64 \\ 3\times 3, 64, G=8 \\ 1\times 1, 256 \end{array} \right] \times 5 \\ \text{SDA}[r=16, L=64], 256 \end{array}$
                             & $\begin{array}{ll} \left[ \begin{array}{llll} 1\times 1, 64 \\ \text{SE}, 64 \\ 1\times 1, 256 \end{array} \right] \times 5 \\ \text{SDA}[r=16, L=64], 256 \end{array}$
                             & $\begin{array}{ll} \left[ \begin{array}{llll} 1\times 1, 64 \\ \text{Res2}, 104 \\ 1\times 1, 256 \end{array} \right] \times 5 \\ \text{SDA}[r=16, L=64], 256 \end{array}$\\
    \specialrule{0.10em}{1.0pt}{1.0pt}
    stage2 & 28$\times$28    & $\begin{array}{ll} \left[ \begin{array}{llll} 1\times 1, 128 \\ 3\times 3, 128 \\ 1\times 1, 512 \end{array} \right] \times 4 \end{array}$
                             & $\begin{array}{ll} \left[ \begin{array}{llll} 1\times 1, 128 \\ 3\times 3, 128, G=8 \\ 1\times 1, 512 \end{array} \right] \times 6 \\ \text{SDA}[r=16, L=64], 512 \end{array}$
                             & $\begin{array}{ll} \left[ \begin{array}{llll} 1\times 1, 128 \\ \text{SE}, 128 \\ 1\times 1, 512 \end{array} \right] \times 6 \\ \text{SDA}[r=16, L=64], 512 \end{array}$
                             & $\begin{array}{ll} \left[ \begin{array}{llll} 1\times 1, 128 \\ \text{Res2}, 208 \\ 1\times 1, 512 \end{array} \right] \times 6 \\ \text{SDA}[r=16, L=64], 512 \end{array}$\\
    \specialrule{0.10em}{1.0pt}{1.0pt}
    stage3 & 14$\times$14    & $\begin{array}{ll} \left[ \begin{array}{llll} 1\times 1, 256 \\ 3\times 3, 256 \\ 1\times 1, 1024 \end{array} \right] \times 6 \end{array}$
                             & $\begin{array}{ll} \left[ \begin{array}{llll} 1\times 1, 256 \\ 3\times 3, 256, G=8 \\ 1\times 1, 1024 \end{array} \right] \times 12 \\ \text{SDA}[r=16, L=64], 1024 \end{array}$
                             & $\begin{array}{ll} \left[ \begin{array}{llll} 1\times 1, 256 \\ \text{SE}, 256 \\ 1\times 1, 1024 \end{array} \right] \times 12 \\ \text{SDA}[r=16, L=64], 1024 \end{array}$
                             & $\begin{array}{ll} \left[ \begin{array}{llll} 1\times 1, 256 \\ \text{Res2}, 416 \\ 1\times 1, 1024 \end{array} \right] \times 12 \\ \text{SDA}[r=16, L=64], 1024 \end{array}$\\
    \specialrule{0.10em}{1.0pt}{1.0pt}
    stage4 & 7$\times$7      & $\begin{array}{ll} \left[ \begin{array}{llll} 1\times 1, 512 \\ 3\times 3, 512 \\ 1\times 1, 2048 \end{array} \right] \times 3 \end{array}$
                             & $\begin{array}{ll} \left[ \begin{array}{llll} 1\times 1, 512 \\ 3\times 3, 512, G=8 \\ 1\times 1, 2048 \end{array} \right] \times 5 \\ \text{SDA}[r=16, L=64], 2048 \end{array}$
                             & $\begin{array}{ll} \left[ \begin{array}{llll} 1\times 1, 512 \\ \text{SE}, 512 \\ 1\times 1, 2048 \end{array} \right] \times 5 \\ \text{SDA}[r=16, L=64], 2048 \end{array}$
                             & $\begin{array}{ll} \left[ \begin{array}{llll} 1\times 1, 512 \\ \text{Res2}, 832 \\ 1\times 1, 2048 \end{array} \right] \times 5 \\ \text{SDA}[r=16, L=64], 2048 \end{array}$\\
    \specialrule{0.10em}{1.0pt}{1.0pt}
    classifier & 1$\times$1  & \multicolumn{4}{c}{7$\times$7 global average pool, BN, 1000-d fc, softmax} \\
    \specialrule{0.10em}{1.0pt}{1.0pt}
    \multicolumn{2}{c|}{FLOPs}     &      4.12G      &      3.88G      &      6.72G      &       6.95G\\
    \specialrule{0.10em}{1.0pt}{1.0pt}
    \multicolumn{2}{c|}{Params}    &      25.56M     &     27.22M      &     49.13M      &      45.26M\\
    \specialrule{0.10em}{1.0pt}{1.0pt}
  \end{tabular}}
\end{table*}

We construct our proposed SDA-Net by stacking multiple SDA modules.
Based on ResNet-50, we adjust the number of blocks in each stage from (3, 4, 6, 3) to (5, 6, 12, 5), and set 8 groups in the second convolution of each block for aligning the FLOPs with ResNet-50.
Thanks to the proposed depth attention mechanism, SDA-Net enables a wide range of effective receptive fields for the flexible multi-scale representation ability.

To be specific, if SDA-Net pays more attention to deep blocks of each stage, then it outputs a high level of semantic features with large receptive fields, which allows it to favor the large-scale objects. On the contrary, if shallow blocks are paid more attention to in each stage, then a low level of features is obtained with small receptive fields, being suitable for learning the small-scale objects.
The original ResNet is viewed as a special case of our SDA-Net, which generates the fixed-scale features for the single-scale objects through the only last block. In contrast, our SDA-Net adaptively learns the dynamic multi-scale contexts to achieve better performance for various scales of objects by leveraging the intermediate features instead of just the features of the last block.
More importantly, because of the newly proposed depth attention dimension, our method is orthogonal to other multi-scale attention methods such as EPSANet, SENet, CBAM, PyConv, and Res2Net. Thus, the proposed SDA method further extends the range of effective receptive fields towards small receptive fields for enhancing the model representation ability, which is of similar complexities to EPSANet-101, SENet-101, CBAM-101, PyConv-101, and Res2Net-101, respectively. A comparison of ResNet-50, SDA-ResNet-86, SDA-SENet-86 and SDA-Res2Net-86 neural architectures can be found in Table~\ref{tab:1}.

In addition to the above attention mechanism, the composition of $\mathbf{Z}$ is also critical for our SDA-Net to generate the features with rich semantic information according to Eqns.~(\ref{eqn:1}) and~(\ref{eqn:5}).
The key lies on the choice of features from the intermediate blocks.
There are several candidate features, including $\mathbf{x}+F(\mathbf{x})$ and $\gamma(\mathbf{x}+F(\mathbf{x}))$, where $\mathbf{x}$ is the input tensor of the blocks and $F(\mathbf{x})$ represents multiple convolutional layers which learn a residual mapping function.
In this paper, we select the output of each block before activation, i.e., $\mathbf{x}+F(\mathbf{x})$, to constitute the sequence of feature map subsets $\mathbf{Z}$ given the information loss from the activation function $\gamma$.
The choice and effect of these intermediate features will be further discussed in the section on Ablation Study.

\section{Experiments}\label{sec:Experiments}

In this section, we conduct extensive experiments to investigate the effectiveness of our SDA-Net on numerous benchmark CV tasks, i.e., image classification, object detection, and instance segmentation.

\subsection{Implementation Details}\label{sec:ImplementationDetails}

To evaluate the effectiveness of our SDA-Net for the image classification task, we carry out experiments on the large-scale ImageNet-1k~\cite{ILSVRC15} dataset, and deploy the typical ResNet-50 as the trunk branch.
Considering the combination with other multi-scale and attention methods, we also deploy several other trunk branches, including EPSANet, SENet, CBAM, PyConv, and Res2Net.
The ImageNet-1K dataset contains 1000 object classes with 1.28M training images for training samples and 50K validation images for testing samples.
We adopt the same data augmentation and hyper-parameter setting as in~\cite{HeZRS16}. Specifically, the training images are resized to 256$\times$256 and then are randomly cropped to 224$\times$224, followed by random horizontal flipping.
All the models are trained from scratch by stochastic gradient descent (SGD) with weight decay 1e-4, momentum 0.9, and mini-batch size 256. The initial learning rate is set to 0.1, and decreased by a factor of 10 every 30 epochs until 120 epochs. During testing, we resize the validation images to 256$\times$256 and then use the center crop of 224$\times$ 224 for evaluation. Label Smoothing~\cite{SzegedyVISW16} is applied to regularize the network.

To evaluate our SDA method on MS COCO~\cite{LinMBHPRDZ14} for other downstream CV tasks, we use multiple detectors such as Faster R-CNN~\cite{RenHGS15} and Mask R-CNN~\cite{HeGDG17}.
The COCO-2017 dataset consists of about 118K training images and 5K validation images as testing ones over 80 foreground object classes and one background class.
These detectors are implemented with MMDetection toolbox~\cite{MMDetection} and are configured with its default settings.
Concretely, during training, the input images are resized such that the short edge is 800 pixels. We train all the models for 12 epochs using SGD with an initial learning rate of 0.02 which is decreased by 10 at the 8th and 11th epochs.
The other hyper-parameters are set as follows: a weight decay of 1e-4, a momentum of 0.9, and a mini-batch size of 16 (4 GPUs with 4 images per GPU).

\subsection{Image Classification}\label{sec:ImageClassification}

We compare our SDA-Net with the SOTA multi-scale networks on ImageNet-1K, including bL-ResNet~\cite{ChenFMSF19}, ScaleNet~\cite{LiKCZ19}, EPSANet~\cite{ZhangZLZM21}, PyConv~\cite{DutaLZS20}, and Res2Net~\cite{GaoCZZYT19}, to evaluate the proposed method on the image classification task. Table~\ref{tab:2} shows the result comparison with the SOTA multi-scale methods and the result of combining with the part of those multi-scale methods.

Specifically, our SDA-Net outperforms other multi-scale networks by a large margin with similar computational complexity. Our method achieves 1.45\%, 0.78\%, 1.27\% and 0.77\% higher performance in terms of top-1 accuracy than bL-ResNet, ScaleNet, EPSANet, PyConv, and Res2Net, respectively. Moreover, the model integrating our method with other multi-scale methods significantly improves the recognition performance, even consistently surpassing their 101-layer variant networks, i.e., EPSANet-101, PyConv-101, and Res2Net-101, with a smaller computational cost.
This fully demonstrates the superiority of adaptive aggregation for different receptive fields.

\begin{table*}[!t]
  \scriptsize
  \caption{Classification Comparison among several SOTA multi-scale methods on ImageNet-1K.} 
  \label{tab:2}
  \centering
  \setlength{\tabcolsep}{1.8mm}{
  \begin{tabular}{l|c|c|c|c|c|c}
    \specialrule{0.10em}{0pt}{0pt}
    Method &
    Backbone Model &
    Multi-scale &
    \makecell*[c]{Params\\ (M)} &
    \makecell*[c]{FLOPs\\ (G)} &
    \makecell*[c]{Top-1\\ (\%)} &
    \makecell*[c]{Top-5\\ (\%)} \\
    \specialrule{0.08em}{0pt}{0.5pt}
    ResNet-50~\cite{HeZRS16}               &  \multirow{7}{*}{ResNet}        & multi-scale &   25.56  &  4.12  &  75.20  &  92.52\\
    bL-ResNet~\cite{ChenFMSF19}            &                                 & multi-scale &   26.69  &  2.85  &  77.31  &  -\\
    ScaleNet~\cite{LiKCZ19}                &                                 & multi-scale &   31.48  &  3.82  &  77.98  &  93.95\\
    EPSANet~\cite{ZhangZLZM21}             &                                 & multi-scale &   22.59  &  3.60  &  77.49  &  93.54\\
    PyConv~\cite{DutaLZS20}                &                                 & multi-scale &   24.85  &  3.88  &  77.88  &  93.80\\
    Res2Net~\cite{GaoCZZYT19}              &                                 & multi-scale &   25.70  &  4.26  &  77.99  &  93.85\\
    SDA-ResNet-86 (ours)                   &                                 & multi-scale &   27.22  &  3.88  &  \textbf{78.76}  &  \textbf{94.37}\\
    \specialrule{0.08em}{0pt}{0.5pt}
    EPSANet-101~\cite{ZhangZLZM21}         &  \multirow{3}{*}{EPSANet}       & multi-scale &   38.90  &  6.82  &  78.43  &  94.11\\
    EPSANet-86 (our impl.)                 &                                 & multi-scale &   36.67  &  5.84  &  77.71  &  93.83\\
    SDA-EPSANet-86 (ours)                  &                                 & multi-scale &   39.45  &  5.85  &  \textbf{78.80}  &  \textbf{94.34}\\
    \specialrule{0.08em}{0pt}{0.5pt}
    PyConv-101~\cite{DutaLZS20}            &  \multirow{3}{*}{PyConv}        & multi-scale &   44.63  &  8.42  &  79.22  &  94.43\\
    PyConv-86 (our impl.)                  &                                 & multi-scale &   40.55  &  6.22  &  78.63  &  94.26\\
    SDA-PyConv-86 (ours)                   &                                 & multi-scale &   43.33  &  6.22  &  \textbf{79.27}  &  \textbf{94.65}\\
    \specialrule{0.08em}{0pt}{0.5pt}
    Res2Net-101~\cite{GaoCZZYT19}          &  \multirow{3}{*}{Res2Net}       & multi-scale &   45.21  &  8.10  &  79.19  &  94.43\\
    Res2Net-86 (our impl.)                 &                                 & multi-scale &   42.47  &  6.94  &  78.63  &  94.21\\
    SDA-Res2Net-86 (ours)                  &                                 & multi-scale &   45.26  &  6.95  &  \textbf{79.36}  &  \textbf{94.70}\\
    \specialrule{0.10em}{0pt}{0pt}
  \end{tabular}}
\end{table*}

To further evaluate the proposed method for the image classification task, our SDA-Net is compared with the SOTA attention methods with different attention dimensions on ImageNet-1K, such as SENet~\cite{HuSS18}, FcaNet~\cite{QinZWL20}, GCNet~\cite{CaoXLWH19}, CBAM~\cite{WooPLS18} and EPSANet~\cite{ZhangZLZM21}.
We present the result compared with the SOTA attention methods and the result of combining with the part of those attention methods in Table~\ref{tab:3}.

In particular, our SDA-Net achieves the best performance among these attention methods with similar complexities. For the channel attention dimension, our method outperforms SENet by more than 2\% in terms of top-1 accuracy. For the spatial-related attention dimension, our SDA-Net obtains a 1.42\% higher gain than CBAM. Our methods boost 1.27\% top-1 accuracy for the branch attention method, i.e., EPSANet. The performance gap between our SDA-Net and FcaNet is the smallest (0.24\%).
Additionally, we observe 0.66\%, 0.69\%, and 1.09\% improvement when combining with the typical networks of different attention dimensions, i.e., SENet, CBAM, and EPSANet, respectively. Our SDA-Net even consistently surpasses their 101-layer network variants with less memory and computation cost.
This fully shows the effectiveness of our proposed depth attention mechanism.

\begin{table*}[!t]
  \scriptsize
  \caption{Classification Comparison among several SOTA attention methods on ImageNet-1K.} 
  \label{tab:3}
  \centering
  \setlength{\tabcolsep}{1.8mm}{
  \begin{tabular}{l|c|c|c|c|c|c}
    \specialrule{0.10em}{0pt}{0pt}
    Method &
    Backbone Model &
    Attention Dimension &
    \makecell*[c]{Params\\ (M)} &
    \makecell*[c]{FLOPs\\ (G)} &
    \makecell*[c]{Top-1\\ (\%)} &
    \makecell*[c]{Top-5\\ (\%)} \\
    \specialrule{0.08em}{0pt}{0.5pt}
    ResNet-50~\cite{HeZRS16}               &  \multirow{13}{*}{ResNet}      & -                     &   25.56  &  4.12  &  75.20  &  92.52\\
    SENet~\cite{HuSS18}                    &                                & channel               &   28.09  &  4.09  &  76.71  &  93.38\\
    ECANet~\cite{WangWZLZH20}              &                                & channel               &   25.56  &  4.09  &  77.48  &  93.68\\
    GSoP~\cite{GaoXWL19}                   &                                & channel               &   28.05  &  6.18  &  77.68  &  93.98\\
    FcaNet~\cite{QinZWL20}                 &                                & channel               &   28.07  &  4.09  &  78.52  &  94.14\\
    A$^2$Net~\cite{ChenKLYF18}             &                                & spatial               &   33.00  &  6.50  &  77.00  &  93.50\\
    AANet~\cite{BelloZVSL19}               &                                & spatial               &   25.80  &  4.15  &  77.70  &  93.80\\
    GCNet~\cite{CaoXLWH19}                 &                                & spatial               &   28.11  &  4.13  &  77.70  &  93.66\\
    BAM~\cite{ParkWLS18}                   &                                & channel+spatial       &   25.92  &  4.17  &  75.98  &  92.82\\
    CBAM~\cite{WooPLS18}                   &                                & channel+spatial       &   30.62  &  4.10  &  77.34  &  93.69\\
    SANet~\cite{ZhangY21}                  &                                & channel+spatial       &   25.56  &  4.09  &  77.72  &  93.80\\
    EPSANet~\cite{ZhangZLZM21}             &                                & branch                &   22.59  &  3.60  &  77.49  &  93.54\\
    SDA-ResNet-86 (ours)                   &                                & depth                 &   27.22  &  3.88  &  \textbf{78.76}  &  \textbf{94.37}\\
    \specialrule{0.08em}{0pt}{0.5pt}
    SENet-101~\cite{HuSS18}                &  \multirow{3}{*}{SENet}        & channel               &   49.33  &  7.81  &  77.62  &  93.93\\
    SENet-86 (our impl.)                   &                                & channel               &   46.35  &  6.71  &  78.29  &  93.92\\
    SDA-SENet-86 (ours)                    &                                & depth+channel         &   49.13  &  6.72  &  \textbf{78.95}  &  \textbf{94.50}\\
    \specialrule{0.08em}{0pt}{0.5pt}
    CBAM-101~\cite{WooPLS18}               &  \multirow{3}{*}{CBAM}         & channel+spatial       &   54.04  &  7.81  &  78.49  &  94.31\\
    CBAM-86 (our impl.)                    &                                & channel+spatial       &   50.75  &  6.71  &  78.36  &  94.09\\
    SDA-CBAM-86 (ours)                     &                                & depth+channel+spatial &   53.53  &  6.72  &  \textbf{79.05}  &  \textbf{94.44}\\
    \specialrule{0.08em}{0pt}{0.5pt}
    EPSANet-101~\cite{ZhangZLZM21}         &  \multirow{3}{*}{EPSANet}      & branch                &   38.90  &  6.82  &  78.43  &  94.11\\
    EPSANet-86 (our impl.)                 &                                & branch                &   36.67  &  5.84  &  77.71  &  93.83\\
    SDA-EPSANet-86 (ours)                  &                                & depth+branch          &   39.45  &  5.85  &  \textbf{78.80}  &  \textbf{94.34}\\
    \specialrule{0.10em}{0pt}{0pt}
  \end{tabular}}
\end{table*}

\subsection{Object Detection}\label{sec:Object_Detection}

We apply our method to the object detection task to explore its generalization ability. We evaluate our SDA-Net as the backbone of Faster R-CNN and Mask R-CNN. All models are trained along with FPN~\cite{LinDGHHB17} and are tested on the MS COCO dataset. Table~\ref{tab:4} shows the object detection results. The SDA-Net based models outperform their counterparts by a significant margin with both faster-RCNN and Mask-RCNN. Compared with ResNet-50, SENet-50, and Res2Net-50, their SDA-Net based models achieve 3.4\%, 3.4\%, and 8.4\% higher AP performance for Faster-RCNN, respectively, and achieve 3.4\%, 2.8\%, and 2.7\% higher AP performance for Mask-RCNN, respectively, surpassing even their 101-layer variant models in terms of almost all indicators. In the Appendix section, some visual results of object detection on challenging examples are illustrated in Fig.~\ref{fig:6}.

\begin{table*}[!t]
  \scriptsize
  \caption{Object detection results of different methods on COCO val 2017.}
  \label{tab:4}
  \centering
  \setlength{\tabcolsep}{1.8mm}{
  \begin{tabular}{l|c|c|c|c|c|c|c|c|c}
    \specialrule{0.10em}{0pt}{0pt}
    Detector &
    Model &
    \makecell*[c]{Params\\ (M)} &
    \makecell*[c]{FLOPs\\ (G)} &
    \makecell*[c]{$AP$\\ (\%)} &
    \makecell*[c]{$AP_{50}$\\ (\%)} &
    \makecell*[c]{$AP_{75}$\\ (\%)} &
    \makecell*[c]{$AP_{S}$\\ (\%)} &
    \makecell*[c]{$AP_{M}$\\ (\%)} &
    \makecell*[c]{$AP_{L}$\\ (\%)} \\
    \specialrule{0.10em}{0pt}{0pt}
    \multirow{11}{*}{Faster-RCNN} &ResNet-50~\cite{HeZRS16}                   &  41.53  &  207.07  & 36.4 & 58.2 & 39.2 & 21.8 & 40.0 & 46.2\\
                                  &EPSANet~\cite{ZhangZLZM21}                 &  38.56  &  197.07  & 39.2 & 60.3 & 42.3 & 22.8 & 42.4 & 51.1\\
                                  &SENet~\cite{HuSS18}                        &  44.02  &  207.18  & 37.7 & 60.1 & 40.9 & 22.9 & 41.9 & 48.2\\
                                  &ECANet~\cite{WangWZLZH20}                  &  41.53  &  207.18  & 38.0 & 60.6 & 40.9 & 23.4 & 42.1 & 48.0\\
                                  &FcaNet~\cite{QinZWL20}                     &  44.02  &  215.63  & 39.0 & \textbf{61.1} & 42.3 & \textbf{23.7} & 42.8 & 49.6\\
                                  &Res2Net~\cite{GaoCZZYT19}                  &  41.67  &  215.52  & 33.7 & 53.6 & -    & 14.0 & 38.3 & 51.1\\
                                  &SDA-ResNet-86 (ours)                       &  43.11  &  202.93  & \textbf{39.8} & 60.7 & \textbf{43.2} & 22.9 & \textbf{43.9} & \textbf{51.4}\\
                                  \cline{2-10}
                                  &SENet-101~\cite{HuSS18}                    &  65.24  &  295.58  & 39.6 & 62.0 & 43.1 & 23.7 & 44.0 & 51.4\\
                                  &SDA-SENet-86 (ours)                        &  64.81  &  260.93  & \textbf{41.0} & \textbf{62.1} & \textbf{44.5} & \textbf{23.9} & \textbf{44.9} & \textbf{53.4}\\
                                  \cline{2-10}
                                  &Res2Net-101~\cite{GaoCZZYT19}              &  61.18  &  293.68  & -    & -    & -    & -    & -    & -\\
                                  &SDA-Res2Net-86 (ours)                      &  60.98  &  265.70  & 42.0 & 62.7 & 45.5 & 24.1 & 45.8 & 55.0\\
    \specialrule{0.08em}{0.5pt}{0.5pt}
    \multirow{11}{*}{Mask-RCNN}   &ResNet-50~\cite{HeZRS16}                   &  44.17  &  261.81  & 37.2 & 58.9 & 40.3 & 22.2 & 40.7 & 48.0\\
                                  &EPSANet~\cite{ZhangZLZM21}                 &  41.20  &  248.53  & 40.0 & 60.9 & 43.3 & 22.3 & 43.2 & 52.8\\
                                  &SENet~\cite{HuSS18}                        &  46.66  &  261.93  & 38.7 & 60.9 & 42.1 & 23.4 & 42.7 & 50.0\\
                                  &ECANet~\cite{WangWZLZH20}                  &  44.17  &  261.93  & 39.0 & 61.3 & 42.1 & \textbf{24.2} & 42.8 & 49.9\\
                                  &FcaNet~\cite{QinZWL20}                     &  46.66  &  261.93  & 40.3 & \textbf{62.0} & 44.1 & 25.2 & 43.9 & 52.0\\
                                  &Res2Net~\cite{GaoCZZYT19}                  &  44.31  &  268.59  & 39.6 & 60.9 & 43.1 & 22.0 & 42.3 & 52.8\\
                                  &SDA-ResNet-86 (ours)                       &  45.75  &  256.01  & \textbf{40.6} & 61.1 & \textbf{44.4} & 23.4 & \textbf{44.3} & \textbf{53.8}\\
                                  \cline{2-10}
                                  &SENet-101~\cite{HuSS18}                    &  67.88  &  348.65  & 40.7 & 62.5 & 44.3 & 23.9 & 45.2 & 52.8\\
                                  &SDA-SENet-86 (ours)                        &  67.46  &  314.00  & \textbf{41.5} & \textbf{62.8} & \textbf{45.2} & \textbf{24.8} & \textbf{45.6} & \textbf{53.7}\\
                                  \cline{2-10}
                                  &Res2Net-101~\cite{GaoCZZYT19}              &  63.82  &  346.74  & 41.8 & 62.6 & 45.6 & 23.4 & 45.5 & 55.6\\
                                  &SDA-Res2Net-86 (ours)                      &  63.62  &  318.78  & \textbf{42.3} & \textbf{62.9} & \textbf{46.4} & \textbf{24.5} & \textbf{46.2} & \textbf{55.9}\\
    \specialrule{0.10em}{0pt}{0pt}
  \end{tabular}}
\end{table*}

\subsection{Instance Segmentation}\label{sec:Instance_Segmentation}

Besides the object detection task, we also evaluate our method on the instance segmentation task to further verify its generalization ability. Similarly, we also use our SDA-Net with FPN~\cite{LinDGHHB17} as the backbone of Mask R-CNN and test their performance on the MS COCO dataset. The performance of instance segmentation is shown in Table~\ref{tab:5}. Compared with other methods, our proposed method achieves the best performance. The improvement of AP are 2.4\%, 1.6\%, and 2.3\% for ResNet-50, SENet-50, and Res2Net-50, respectively, surpassing even their 101-layer variant models in terms of almost all indicators. In the Appendix section, some visual results of instance segmentation on challenging examples are illustrated in Fig.~\ref{fig:7}.

\begin{table*}[!t]
  \scriptsize
  \caption{Instance segmentation results of different methods on COCO val 2017.}
  \label{tab:5}
  \centering
  \setlength{\tabcolsep}{1.8mm}{
  \begin{tabular}{l|c|c|c|c|c|c|c|c|c}
    \specialrule{0.10em}{0pt}{0pt}
    Detector &
    Model &
    \makecell*[c]{Params\\ (M)} &
    \makecell*[c]{FLOPs\\ (G)} &
    \makecell*[c]{$AP$\\ (\%)} &
    \makecell*[c]{$AP_{50}$\\ (\%)} &
    \makecell*[c]{$AP_{75}$\\ (\%)} &
    \makecell*[c]{$AP_{S}$\\ (\%)} &
    \makecell*[c]{$AP_{M}$\\ (\%)} &
    \makecell*[c]{$AP_{L}$\\ (\%)} \\
    \specialrule{0.10em}{0pt}{0pt}
    \multirow{11}{*}{Mask-RCNN}   &ResNet-50~\cite{HeZRS16}                   &  44.17  &  261.81  & 34.1 & 55.5 & 36.2 & 16.1 & 36.7 & 50.0\\
                                  &EPSANet~\cite{ZhangZLZM21}                 &  41.20  &  248.53  & 35.9 & 57.7 & 38.1 & \textbf{18.5} & 38.8 & 49.2\\
                                  &SENet~\cite{HuSS18}                        &  46.66  &  261.93  & 35.4 & 57.4 & 37.8 & 17.1 & 38.6 & 51.8\\
                                  &ECANet~\cite{WangWZLZH20}                  &  44.17  &  261.93  & 35.6 & 58.1 & 37.7 & 17.6 & 39.0 & 51.8\\
                                  &FcaNet~\cite{QinZWL20}                     &  46.66  &  261.93  & 36.2 & \textbf{58.6} & 38.1 & -    & -    & -\\
                                  &Res2Net~\cite{GaoCZZYT19}                  &  44.31  &  268.59  & 35.6 & 57.6 & 37.6 & 15.7 & 37.9 & \textbf{53.7}\\
                                  &SDA-ResNet-86 (ours)                       &  45.75  &  256.01  & \textbf{36.5} & 58.2 & \textbf{39.0} & 16.9 & \textbf{39.5} & 53.5\\
                                  \cline{2-10}
                                  &SENet-101~\cite{HuSS18}                    &  67.88  &  348.65  & 36.8 & 59.3 & 39.2 & 17.2 & 40.3 & \textbf{53.6}\\
                                  &SDA-SENet-86 (ours)                        &  67.46  &  314.00  & \textbf{37.0} & \textbf{59.6} & \textbf{39.3} & \textbf{18.2} & \textbf{40.6} & 53.4\\
                                  \cline{2-10}
                                  &Res2Net-101~\cite{GaoCZZYT19}              &  63.82  &  346.74  & 37.1 & 59.4 & 39.4 & 16.6 & 40.0 & \textbf{55.6}\\
                                  &SDA-Res2Net-86 (ours)                      &  63.62  &  318.78  & \textbf{37.9} & \textbf{59.8} & \textbf{40.7} & \textbf{18.3} & \textbf{41.0} & 55.2\\
    \specialrule{0.10em}{0pt}{0pt}
  \end{tabular}}
\end{table*}

\section{Ablation Study}\label{sec:Ablation_Study}

As discussed in Section~\ref{sec:Methodology}, there are two major factors affecting the multi-scale information flows across stages: the feature sequence in trunk branches and the adaptive depth attention mechanism.

\subsection{The Effect of the Feature Sequence in Trunk Branch}\label{sec:EffectofdifferentfeaturesZintrunkbranch}

We investigate the effect of the feature sequence in trunk branches. Based on Eqn.~(\ref{eqn:5}), we find that the hierarchical features are essential factors to affect the information flow across stages. Empirically, we select two crucial feature elements, i.e., $\gamma(\mathbf{x}+F(\mathbf{x}))$ and $\mathbf{x}+F(\mathbf{x})$, which contain rich semantic information with identity features.

The comparison result is shown in Table~\ref{tab:6}. It is easily seen that $\mathbf{x}+F(\mathbf{x})$ achieves higher accuracy than $\gamma(\mathbf{x}+F(\mathbf{x}))$. We argue that this is because the ReLU function results in the information loss due to zeroing the negative activations, and their combination further increases the loss potential. Therefore, we select $\mathbf{x}+F(\mathbf{x})$ as the component of the feature sequence $\mathbf{Z}$ in trunk branches for better performance.

\begin{table}[!t]
  \scriptsize
  \caption{The result of using different features $Z$ in trunk branches on ImageNet-1K.}
  \label{tab:6}
  \centering
  \setlength{\tabcolsep}{1.8mm}{
  \begin{tabular}{l|l|l|c|c}
    \specialrule{0.10em}{0pt}{0pt}
    Model &
    \makecell*[c]{Params\\ (M)} &
    \makecell*[c]{FLOPs\\ (G)} &
    \makecell*[c]{Top-1\\ (\%)} &
    \makecell*[c]{Top-5\\ (\%)} \\
    \specialrule{0.10em}{0pt}{0pt}
    ResNet-86 (our impl.)                               &   24.44  &  3.87   &  77.43           &  93.60\\
    SDA-ResNet-86 \& $\gamma(\mathbf{x}+F(\mathbf{x}))$ &   27.22  &  3.88   &  78.53           &  94.15\\
    SDA-ResNet-86 \& $\mathbf{x}+F(\mathbf{x})$ (ours)  &   27.22  &  3.88   &  \textbf{78.76}  &  \textbf{94.35}\\
    \specialrule{0.10em}{0pt}{0pt}
  \end{tabular}}
\end{table}

\subsection{The Effect of Adaptive Attention vs. Non-adaptive Attention}\label{sec:Adaptiveattentionvsnon-adaptiveattention}

\begin{table}[!t]
  \scriptsize
  \caption{Comparison between adaptive attention vs. non-adaptive attention on ImageNet-1K.}
  \label{tab:7}
  \centering
  \setlength{\tabcolsep}{1.8mm}{
  \begin{tabular}{l|l|l|c|c}
    \specialrule{0.10em}{0pt}{0pt}
    Model &
    \makecell*[c]{Params\\ (M)} &
    \makecell*[c]{FLOPs\\ (G)} &
    \makecell*[c]{Top-1\\ (\%)} &
    \makecell*[c]{Top-5\\ (\%)} \\
    \specialrule{0.10em}{0pt}{0pt}
    ResNet-86 (our impl.)                               &   24.44  &  3.87   &  77.43           &  93.60\\
    SDA-ResNet-86 \& non-adaptive                       &   24.44  &  3.87   &  78.01           &  94.02\\
    SDA-ResNet-86 \& adaptive (ours)                    &   27.22  &  3.88   &  \textbf{78.76}  &  \textbf{94.35}\\
    \specialrule{0.10em}{0pt}{0pt}
  \end{tabular}}
\end{table}

Next, we compare our adaptive attention with non-adaptive attention to investigate the effect of our depth attention mechanism. These two types of networks have similar architectures except for the part of attention branches. Different from our adaptive SDA-Net, the non-adaptive attention network aggregates the features from different blocks by the simple summation, i.e., $\mathbf{O} = \gamma(\frac{1}{m}\sum_{i=1}^m \mathbf{Z}_i)$, instead of the adaptive weighting operation, and the features of each block are equally important for the multi-scale feature representations.

As shown in Table~\ref{tab:7}, we can see that using our adaptive depth attention method has better performance, achieving close to 0.8\% higher accuracy than the non-adaptive attention competitor, which verifies the effectiveness of our depth attention mechanism. More interestingly, the non-adaptive attention network achieves better performance than the ResNet-86 model, improving the accuracy from 77.43\% to 78.01\%. Therefore, the result fully verifies the success of our SDA-Net design that exploits the hierarchical feature representation of CNNs and the depth attention mechanism for richer information fusion.

\section{Analysis and Interpretation}\label{sec:Analysis_and_Interpretation}

\subsection{Computational Complexity}\label{sec:Computational_Complexity}
In this part, we consider the additional parameters introduced by the proposed SDA module. These additional parameters come from the two 1$\times$1 convolution layers of the attention branch.
Thus, the total number of additional parameters equals the additional computational complexity, which can be calculated by

\begin{equation}
\label{eqn:6}
\begin{aligned}
   \frac{1}{r} \sum_{i=1}^n (m_i + 1)c_i^2,
\end{aligned}
\end{equation}
\noindent
where $n$ is the number of stages, $r$ refers to the reduction ratio, $m_i$ denotes to the number of blocks and $c_i$ denotes the number of output channels for the $i$-th stage.

Our SDA-Net offers a good trade-off between improved accuracy and increased computation complexity. To illustrate the model complexity, we make a comparison between ResNet-86 and SDA-ResNet-86. For a fair comparison, we use the same crop size of 224 for input images. ResNet-86 requires $\sim$24.44M Params (the number of parameters) and $\sim$3.87G FLOPs (floating-point of operations) for a single input image. In comparison, our SDA-ResNet-86 requires $\sim$27.22M Params and $\sim$3.88G FLOPs, corresponding to an acceptable increase in the number of parameters and a slight increase in computation complexity.
At the cost of acceptable additional model complexity, SDA-ResNet-86 significantly surpasses ResNet-101 and ResNet-152 as well as ResNet-50, achieving 3.56\%, 1.93\% and 1.18\% higher top-1 accuracy, respectively.

\subsection{The Range of Effective Receptive Field}\label{sec:Adaptive_ERF_Range}

\begin{figure}[!t]
  \centering
  \subfigure[]{
  \includegraphics[trim=5mm 0mm -5mm 0mm, width=1.20in]{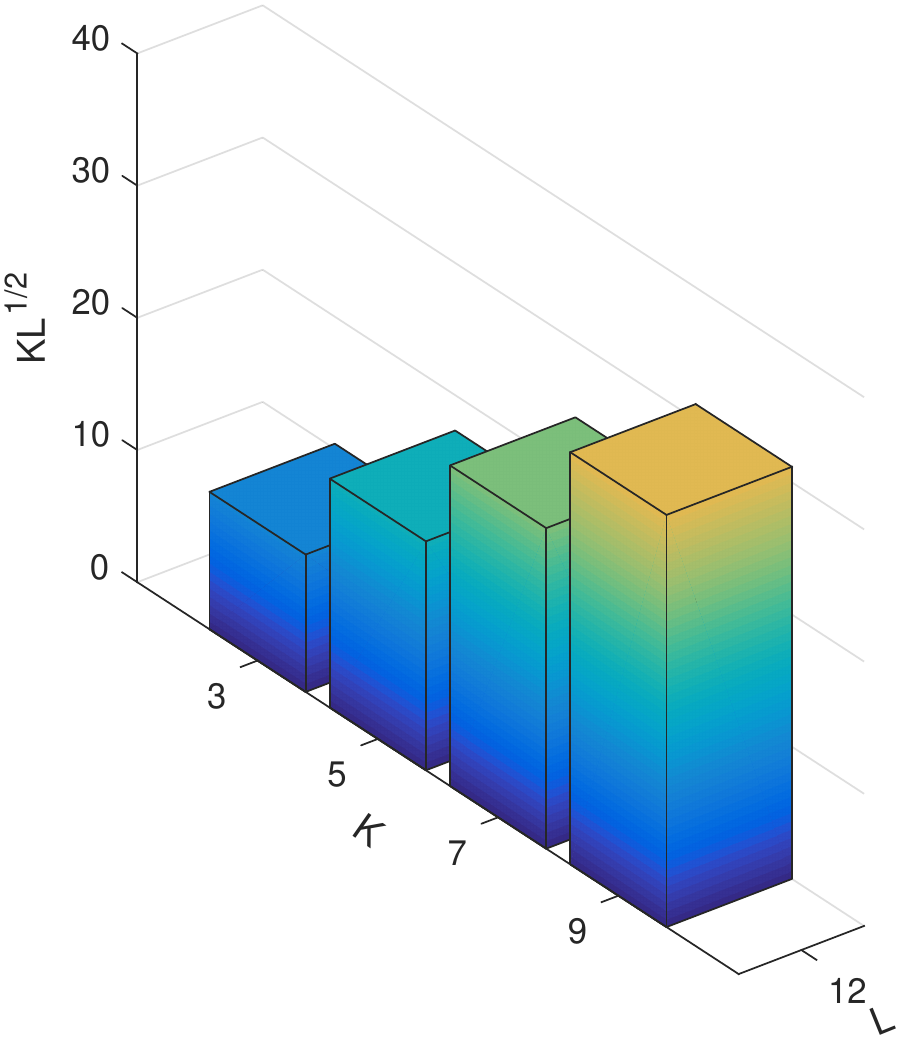}}
  \subfigure[]{
  \includegraphics[trim=5mm 0mm -5mm 0mm, width=1.60in]{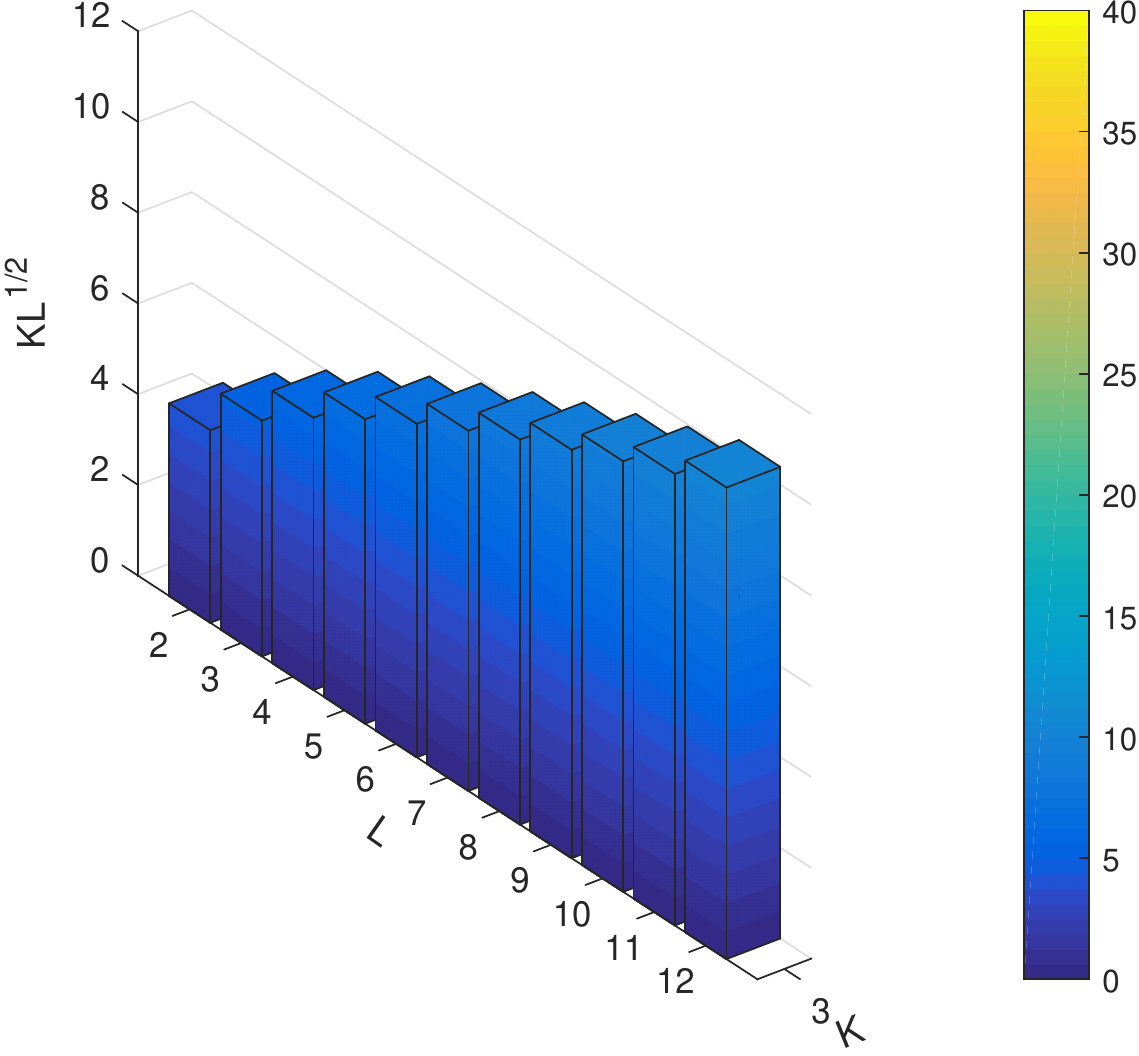}}
  {\caption{The range of adaptive effective receptive field in the second stage. (a) EPSANet. (b) Our SDA-ResNet-86.}
  \label{fig:3}}
\end{figure}

We have demonstrated that the range of the effective receptive field is a crucial issue for the objects at all scales in many CV tasks. For simplicity, take EPSANet as an example, showing the essential difference with our proposed SDA-Net through comparative analysis. According to the theory of effective receptive field, the effective receptive field increases linearly with the kernel size $K$ and sub-linearly with the depth $L$ of networks~\cite{LuoLUZ16}. In other words, the effective receptive field is proportional to $\mathcal{O}(K\sqrt{L})$, which theoretically proves that our proposed method effectively extends the range of effective receptive fields towards small receptive fields because it aggregates shallow features.

In Fig.~\ref{fig:3}, we show the range of effective receptive fields in the second stage for EPSANet-86 and our SDA-ResNet-86, taking only two factors, $K$, and $L$, into consideration. The output of these two networks contains multiple feature components with different receptive fields, in which the large receptive field contributes to capturing semantic information for large-scale objects, and the small receptive field helps to extract local details for small-scale patterns. For EPSANet, it is easily seen that the smallest and largest components of the receptive fields depend on the branch of kernel sizes 3 and 9, respectively.
By contrast, the largest receptive field of our SDA-ResNet-86 is dependent on kernel size 3, which is equivalent to the smallest one of EPSANet, and the other receptive fields gradually shrink as the network depth decreases. Therefore, it is worth noting that our method seeks to strengthen the power of multi-scale representations towards small receptive fields, and the combination with other methods further extends the range of effective receptive fields to achieve stronger representational power.

\subsection{Attention Distribution}\label{sec:Attention_distribution}

\begin{figure}[!t]
  \centering
  \subfigure[]{
  \includegraphics[trim=5mm 0mm -5mm 0mm, width=1.6in]{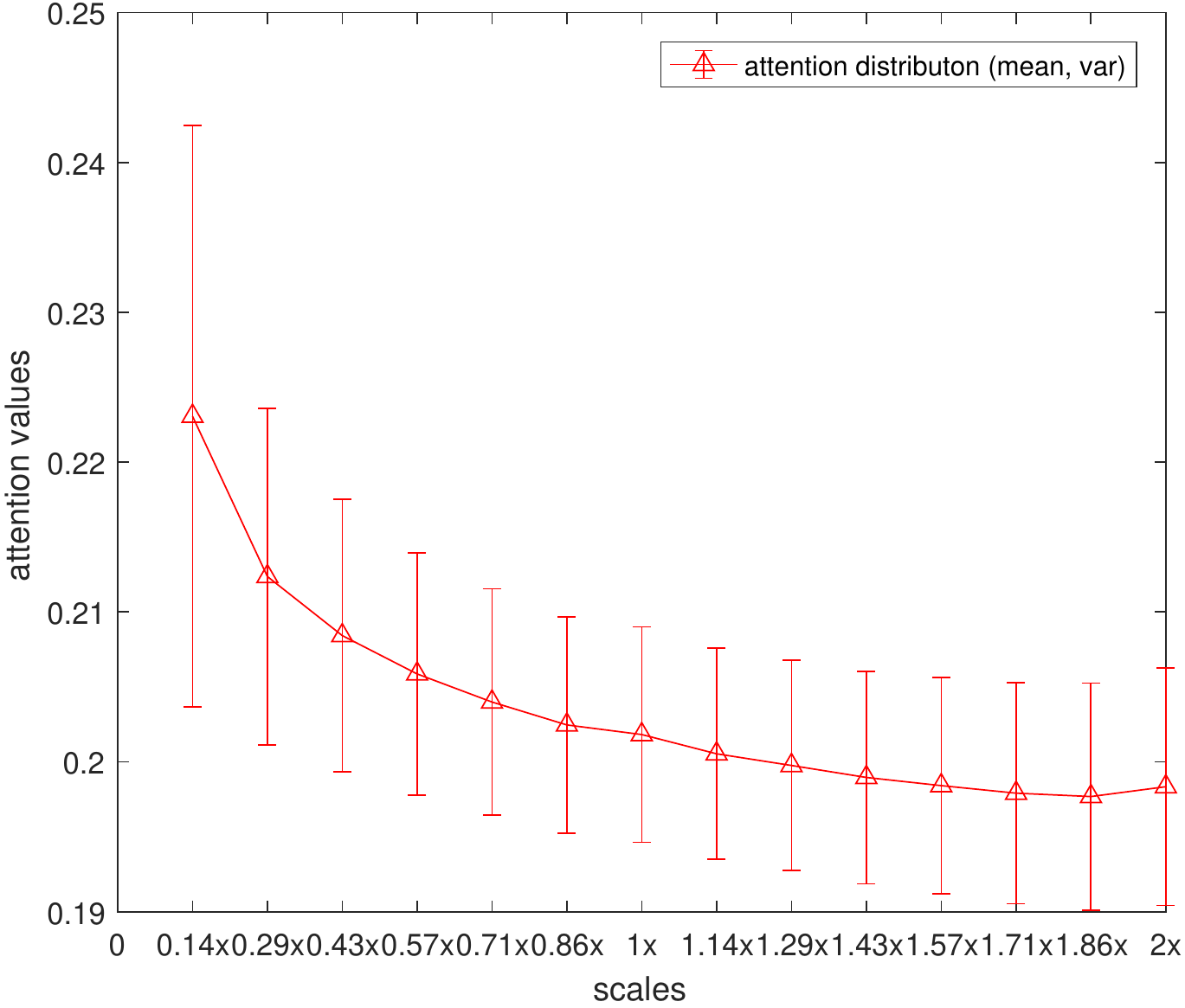}}
  \hspace{0in}
  \subfigure[]{
  \includegraphics[trim=5mm 0mm -5mm 0mm, width=1.6in]{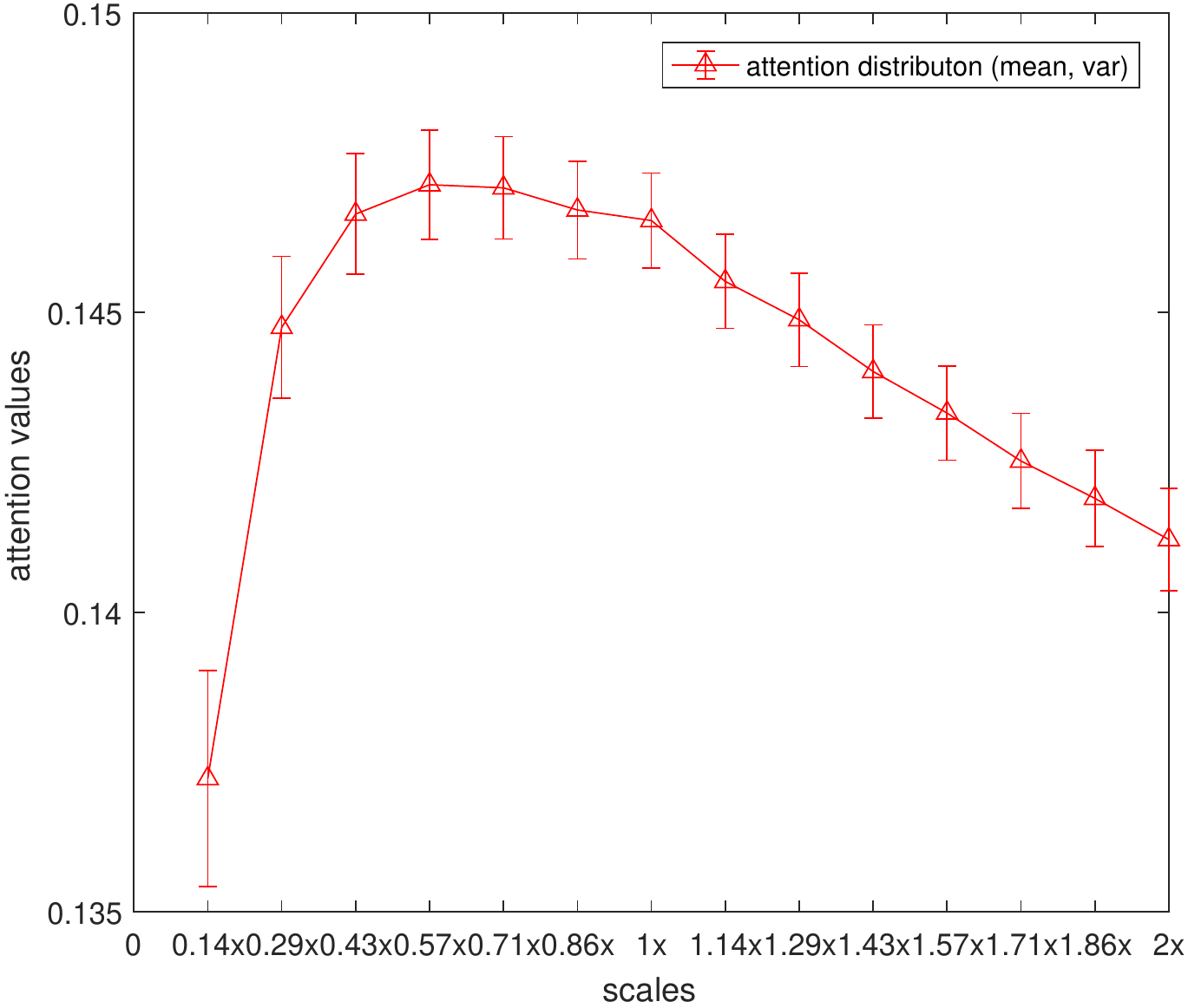}}

  \subfigure[]{
  \includegraphics[trim=5mm 0mm -5mm 0mm, width=1.6in]{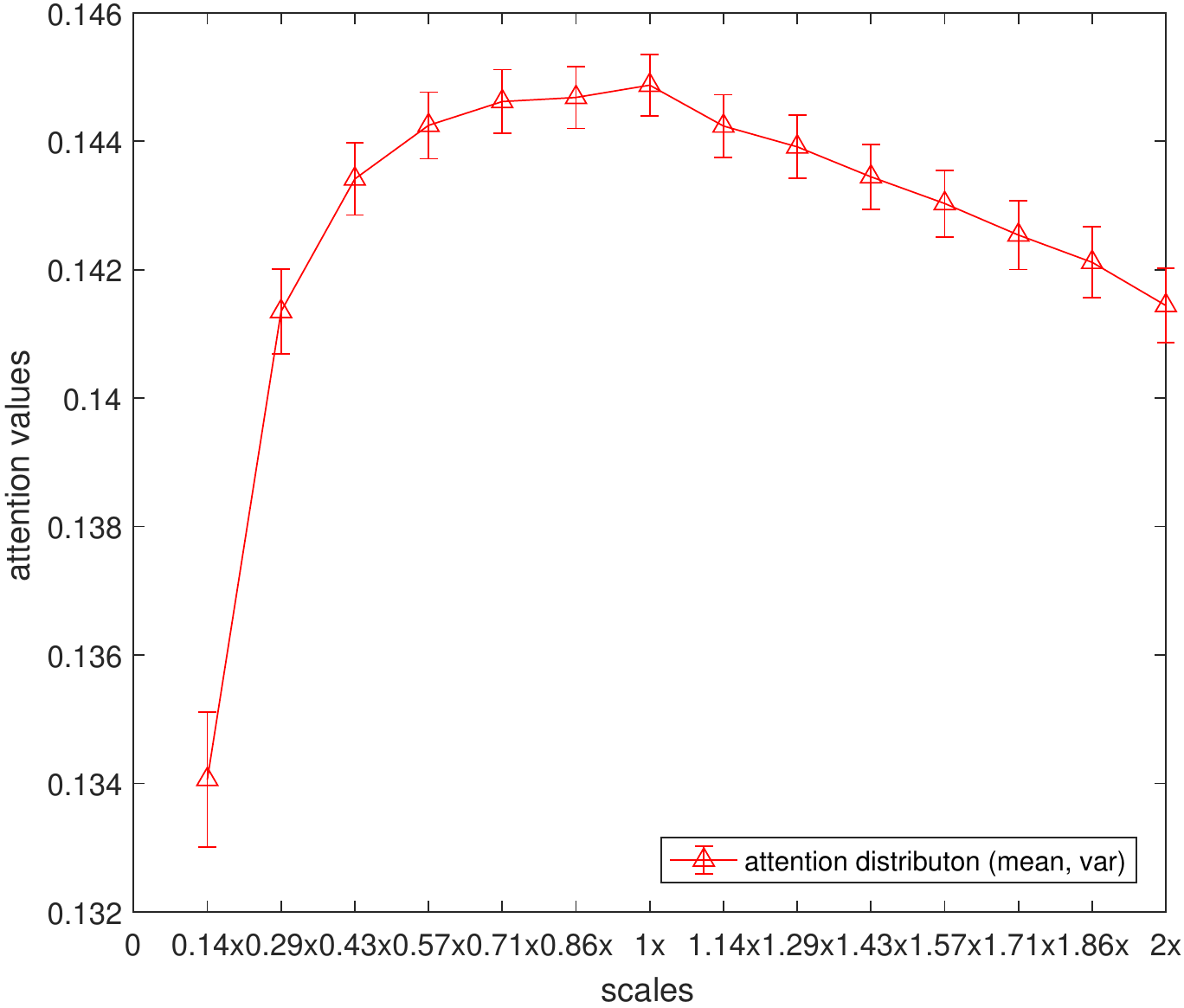}}
  \hspace{0in}
  \subfigure[]{
  \includegraphics[trim=5mm 0mm -5mm 0mm, width=1.6in]{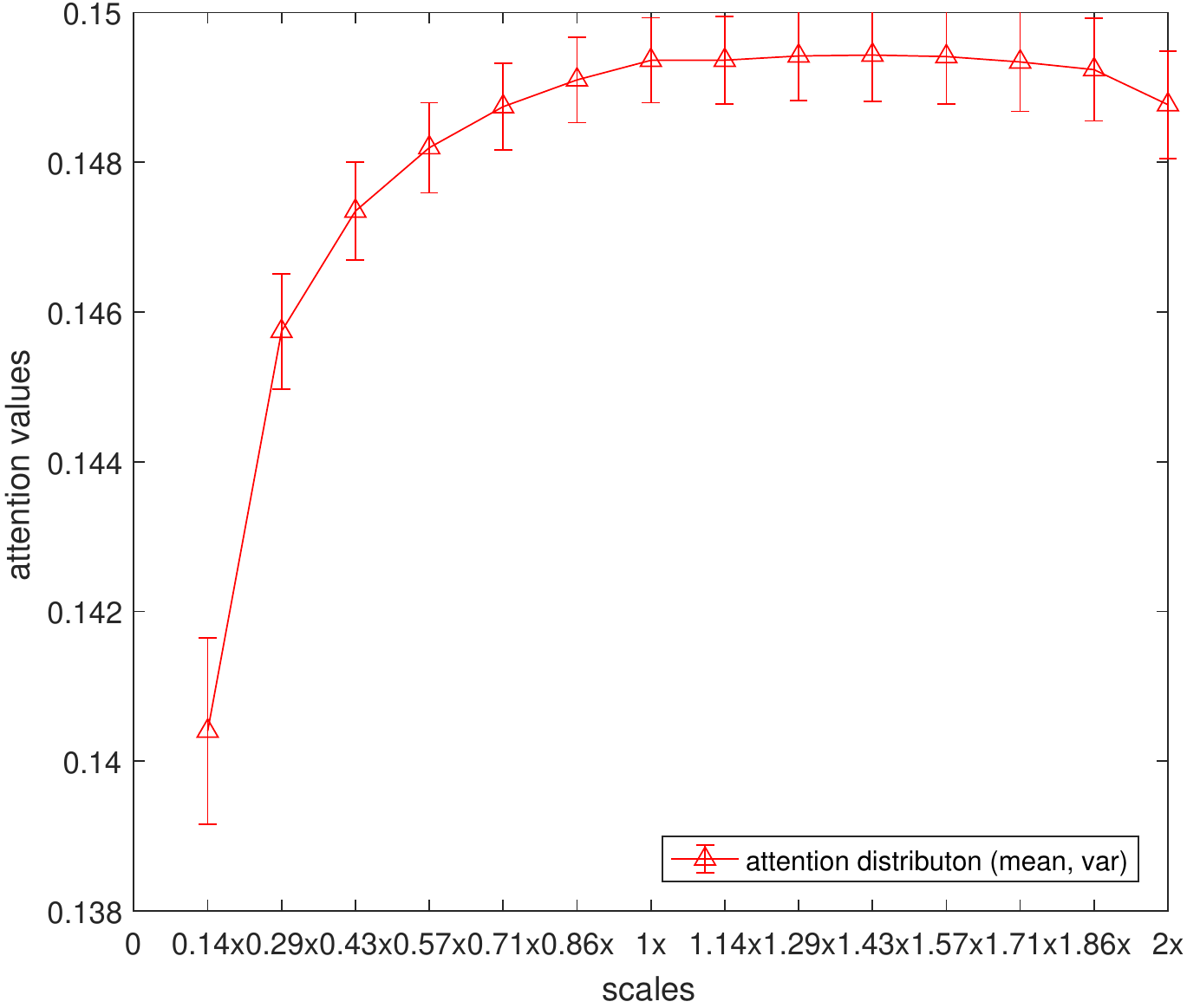}}

  \subfigure[]{
  \includegraphics[trim=5mm 0mm -5mm 0mm, width=1.6in]{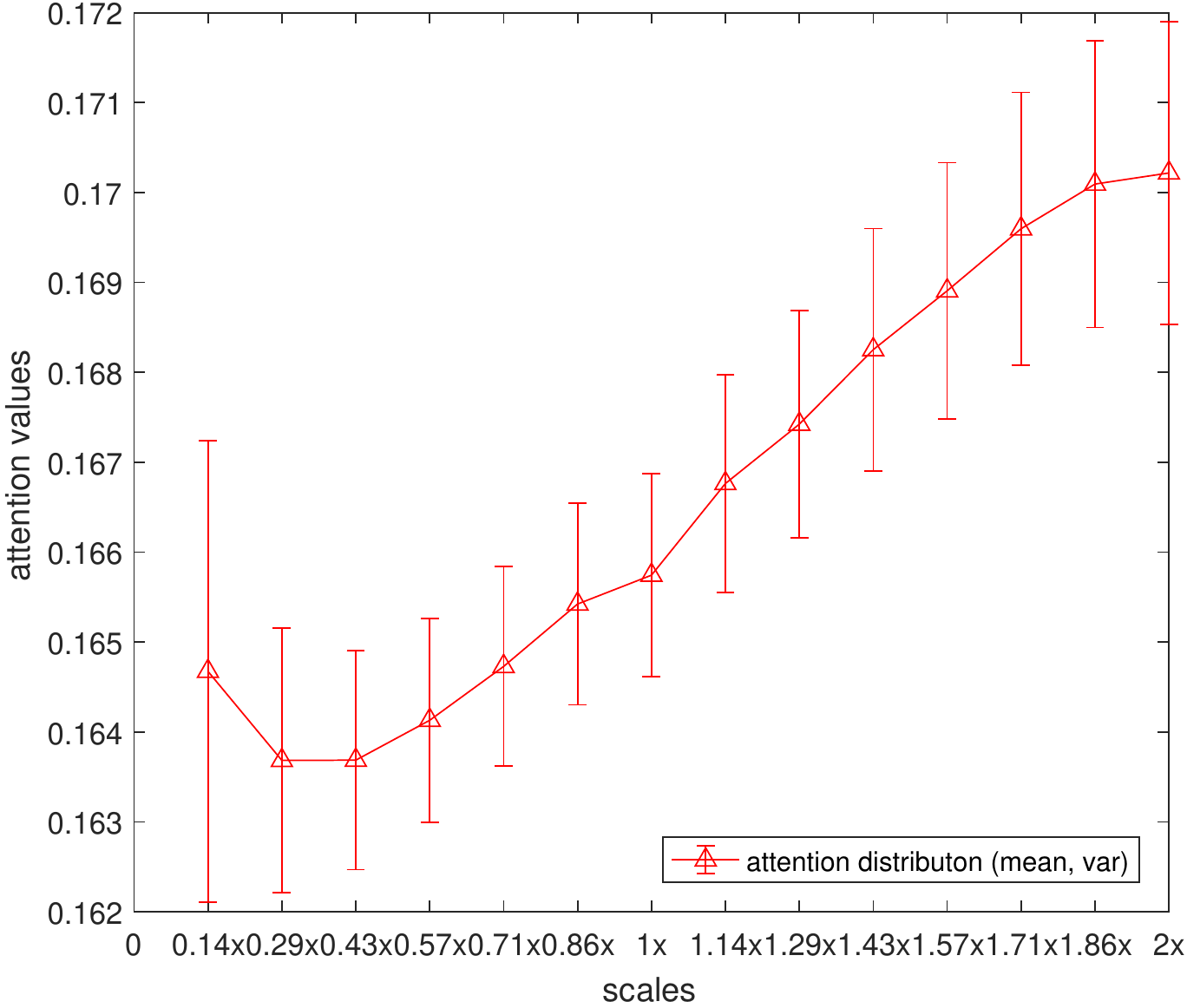}}
  \hspace{0in}
  \subfigure[]{
  \includegraphics[trim=5mm 0mm -5mm 0mm, width=1.65in]{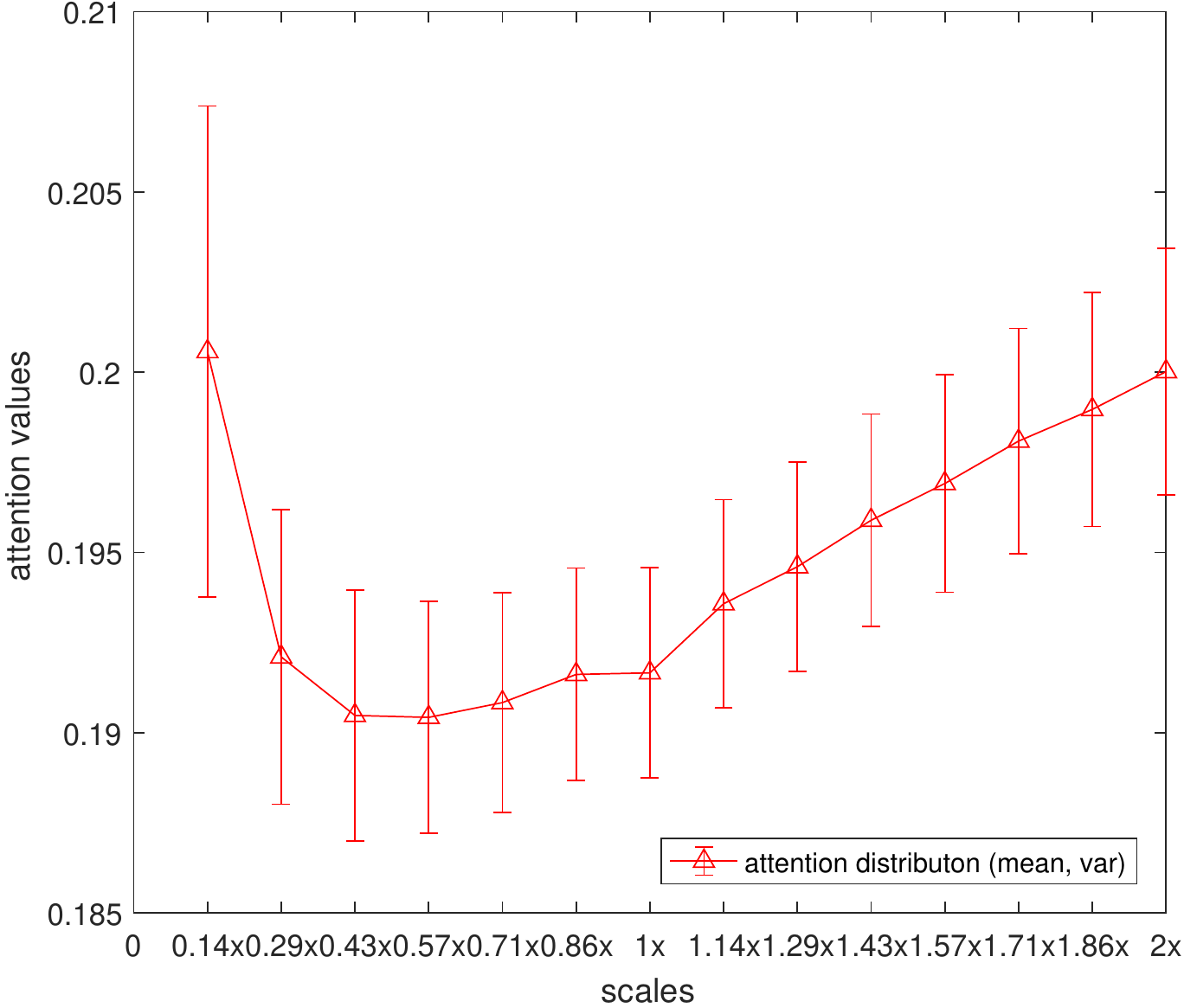}}
  {\caption{Attention distribution (mean, variance) of each block in the second stage which consists of six blocks. (a) The first block. (b) The second block. (c) The third block. (d) The fourth block. (e) The fifth block. (f) The last block.}
  \label{fig:4}}
\end{figure}

To understand how our SDA-Net works for adaptive multi-scale learning by dynamically adjusting receptive fields, we analyze the attention distribution for the same input objects but with different scales. Concretely, we enlarge all the samples of the ImageNet validation set from 0.14$\times$ to 2.0$\times$, followed by a 224$\times$224 central cropping. For the validation set with different scales, we record the attention values of each block and analyze their distributions to mathematically interpret the adaptive adjustment of receptive fields.

Fig.~\ref{fig:4} shows the attention distributions of each block in the second stage. On one hand, it is seen that each block is not equal to be attended for the multi-scale feature representations. On the other hand, as the input objects enlarge, the attention means gradually reduces for the first block; on the contrary, for the fourth block, the attention means tends to increase. For the third block, when the enlargement factor is less than 1$\times$, the attention means dramatically increases. At the 1$\times$ scale, the attention means reaches its peak, and it gradually decreases when the enlargement factor is greater than the 1$\times$ scale. The second block follows a similar pattern to the third block. The fifth and last block has an approximately opposite trend to the second and third blocks. As a result, it suggests that the larger the input objects are, the more attention will be assigned to the features of the deeper blocks, which fully verifies that our SDA-Net adaptively adjusts the receptive fields according to the object scale sizes.

\subsection{Class Activation Mapping}\label{sec:CAM}

To further understand the adaptive multi-scale representation ability of our SDA-Net, we visualize the class activation mapping (CAM) by applying Grad-CAM~\cite{SelvarajuCDVPB2017} to the images from the ImageNet validation set. Grad-CAM is a commonly used visualization method to localize the class-specific discriminative regions.

\begin{figure*}[t]
\begin{minipage}{1\textwidth}
  \centering
  \includegraphics[trim=0mm 5mm 0mm 5mm, width=7.0in]{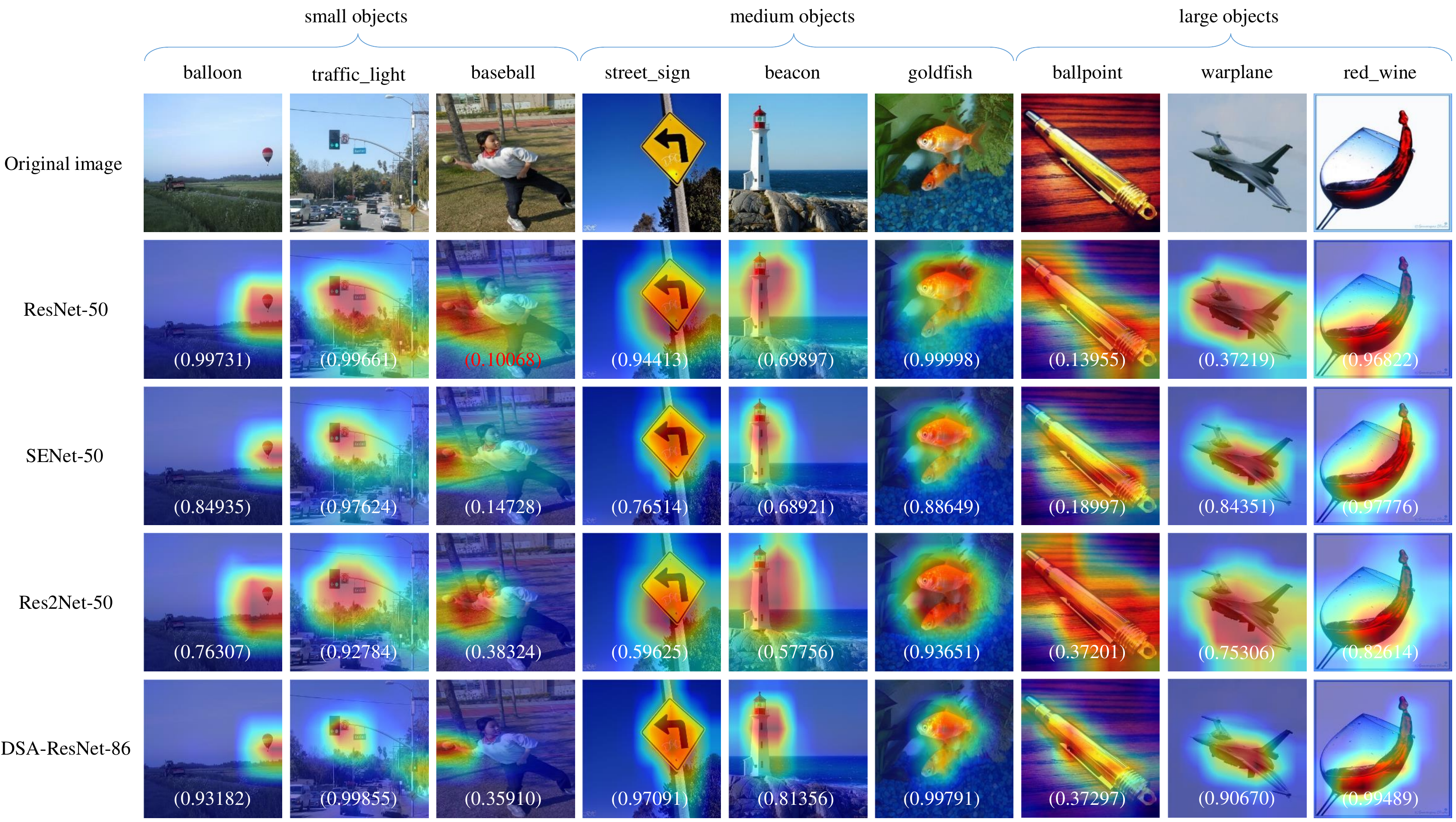}
  \caption{Grad-CAM visualization results. The softmax scores for the target class are displayed at the bottom of each image.}
  \label{fig:5}
\end{minipage}
\end{figure*}

Fig.~\ref{fig:5} shows the visualization results of CAM using ResNet, SENet, Res2Net, and SDA-Net as the backbone. We randomly sample the images with three different size objects~(small, medium, and large)~from the ImageNet validation set.
For small target objects such as `balloon', `laddybug', and  `baseball', the SDA-Net based CAM regions tend to be more precise than other networks. Similarly, SDA-Net provides more precise CAM localizations on medium target objects, such as `street\_sign', `beacon', and `goldfish'. Our SDA-Net covers almost all parts of the large target objects, such as `ballpoint', `bulbul' and `red\_wine', than other networks which cover so much irrelevant context, especially Res2Net. Accordingly, the target class scores of our SDA-Net also tend to be larger. In the Appendix section, more CAM visualization results are illustrated for different methods in Fig.~\ref{fig:8}. As can be seen, the proposed SDA-Net more precisely localizes the object regions with various scale sizes than other networks.

\section{Conclusion And Future Work}\label{sec:Conclusion}
A new attention dimension, i.e., depth, is the first to be introduced, which is an essential factor in addition to existing attention dimensions such as channel, spatial, and branch. Based on this, we present a simple yet effective SDA-Net to explore the depth attention mechanism for the multi-scale feature representations. The network exploits the inherent hierarchical features of CNNs and the long-range block dependency to adaptively adjust the effective receptive fields for different scale-size objects. The proposed method is orthogonal to other multi-scale and attention methods, thus SDA-Net can be integrated with those networks to extend the range of effective receptive fields towards small receptive fields for better performances.
We have achieved state-of-the-art performance in a broad range of CV tasks, including image classification, object detection, and instance segmentation.

We hope that the proposed SDA mechanism will promote further research on attention networks. There are lots of potential improvements, such as designing more efficient attention structures, adopting stronger trunk branches, and combining with the transformer architecture, in future work.

\section*{Acknowledgment}
This work is supported by the National Key R\&D Program of China (Grant No. 2018YFB1004901), by the National Natural Science Foundation of China (Grant No.61672265, U1836218, 62006097), by the 111 Project of Ministry of Education of China (Grant No. B12018), by UK EPSRC GRANT EP/N007743/1, MURI/EPSRC/DSTL GRANT EP/R018456/1, by the Science and Technology Program of University of Jinan (Grant No. XKY1913, XKY1915), and in part by the Natural Science Foundation of Jiangsu Province (Grant No. BK20200593).


\bibliographystyle{IEEEtran}
\bibliography{mydeeplib}


\begin{IEEEbiography}[{\includegraphics[width=1in,height=1.25in,clip,keepaspectratio]{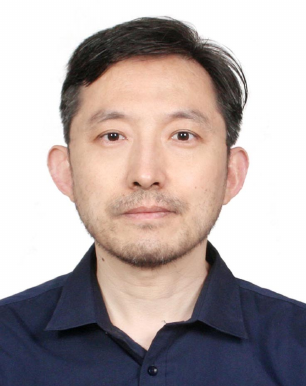}}]{Qingbei Guo}
received the M.S. degree from the School of Computer Science and Technology, Shandong University, Jinan, China, in 2006. He is a member of the Shandong Provincial Key Laboratory of Network based Intelligent Computing and the associate professor in the School of Information Science and Engineering, University of Jinan. He is now a Ph.D. student at Jiangnan University, Wuxi, China. His current research interests include wireless sensor networks, deep learning/machine learning, computer vision and neuron networks.
\end{IEEEbiography}

\begin{IEEEbiography}[{\includegraphics[width=1in,height=1.25in,clip,keepaspectratio]{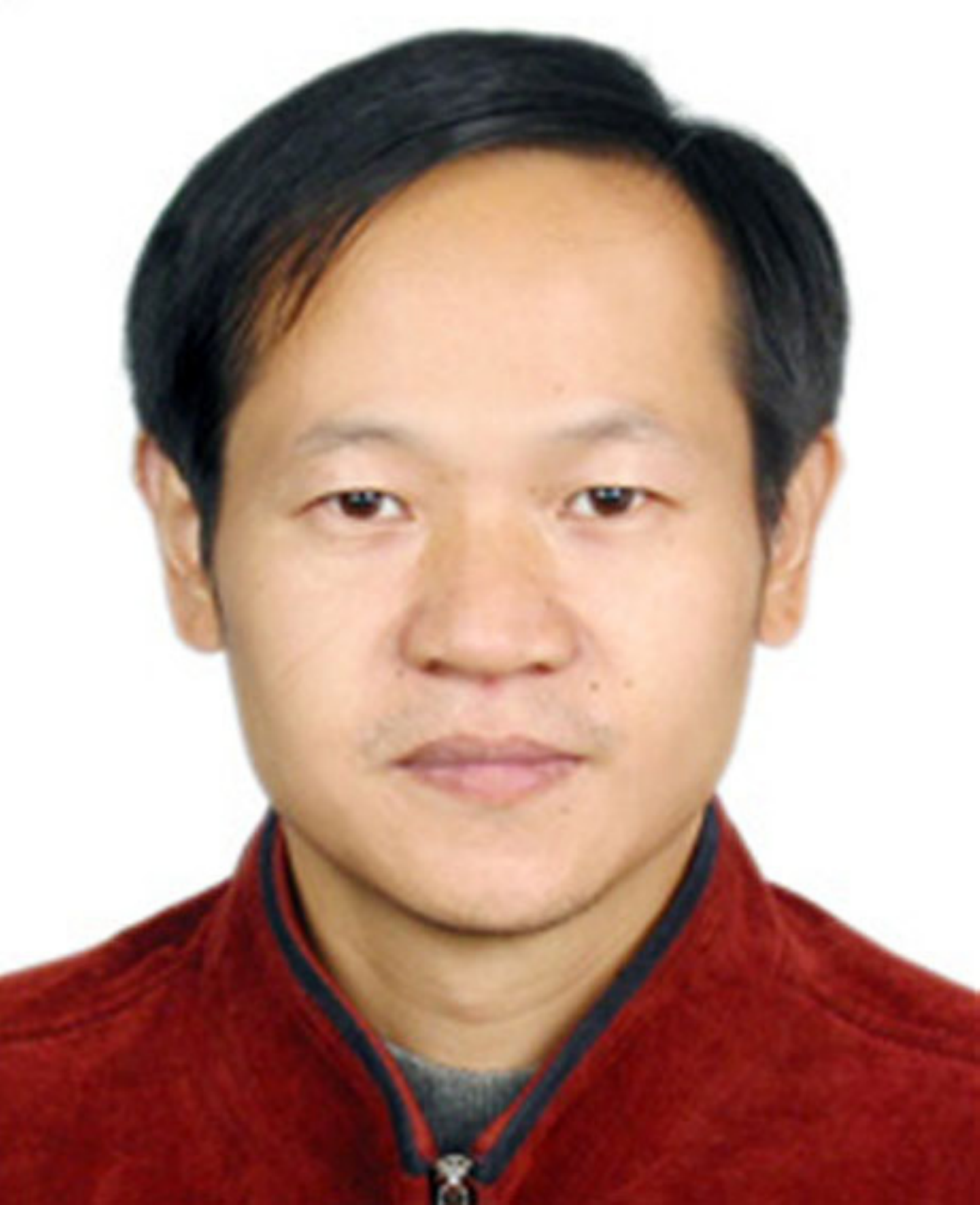}}]{Xiao-jun Wu}
received his B.S. degree in mathematics from Nanjing Normal University, Nanjing, PR China in 1991 and M.S. degree in 1996, and Ph.D. degree in Pattern Recognition and Intelligent System in 2002, both from Nanjing University of Science and Technology, Nanjing, PR China, respectively. He was a fellow of United Nations University, International Institute for Software Technology (UNU/IIST) from 1999 to 2000. From 1996 to 2006, he taught in the School of Electronics and Information, Jiangsu University of Science and Technology where he was an exceptionally promoted professor. He joined the School of Information Engineering, Jiangnan University in 2006 where he is a professor. He won the most outstanding postgraduate award by Nanjing University of Science and Technology. He has published more than 300 papers in his fields of research. He was a visiting researcher in the Centre for Vision, Speech, and Signal Processing (CVSSP), University of Surrey, UK from 2003 to 2004. His current research interests are pattern recognition, computer vision, fuzzy systems, neural networks and intelligent systems.
\end{IEEEbiography}


\begin{IEEEbiography}[{\includegraphics[width=1in,height=1.25in,clip,keepaspectratio]{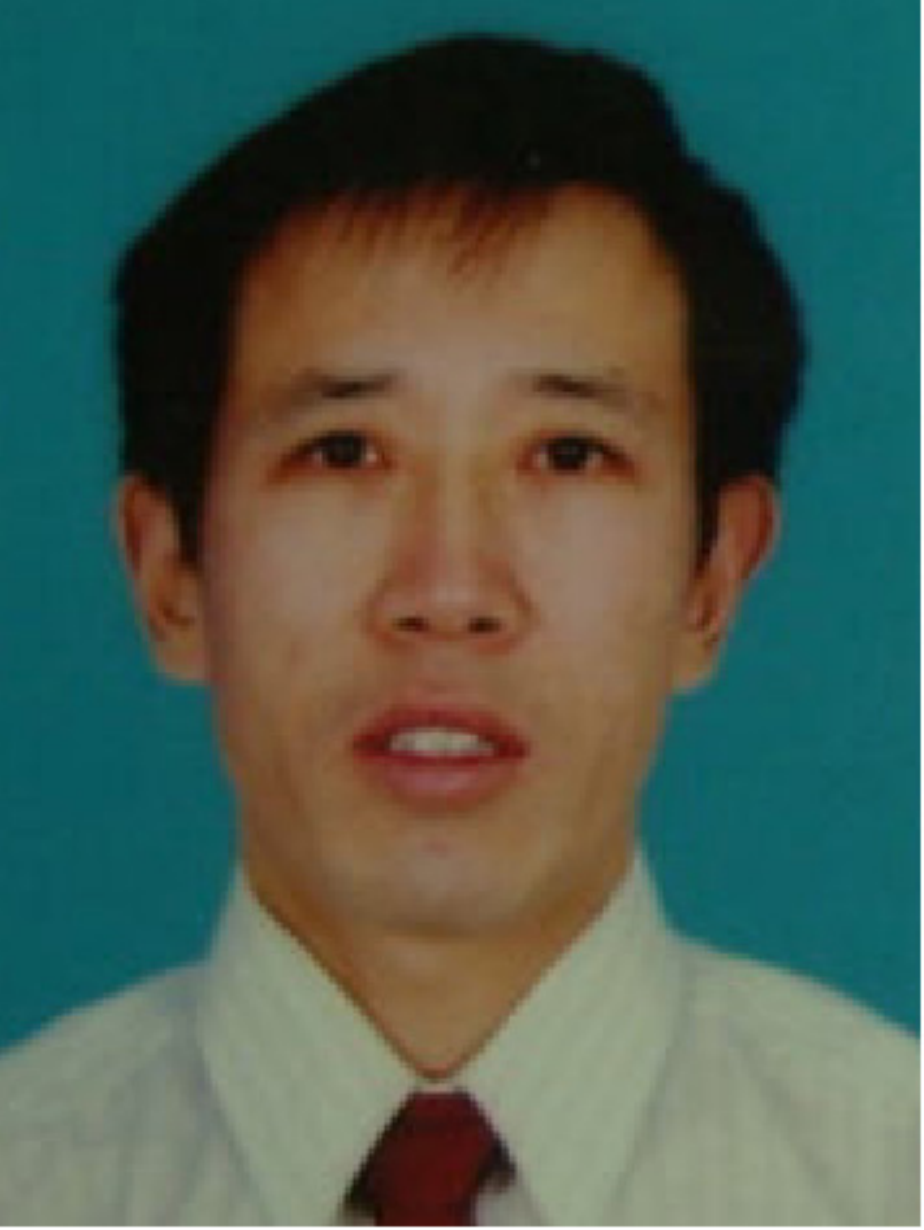}}]{Zhiquan Feng}
received his M.S. degree in Computer Software from Northwestern Polytechnical University, China, in 1995, and Ph.D. degree in Computer Science \& Engineering from Shandong University, China, in 2006. He is currently a Professor at University of Jinan, China. Dr. Feng is a visiting Professor of Sichuang Mianyang Normal University. As the first author or corresponding author he has published more than 100 papers in international journals and conference proceedings, 2 books, and 30 patents in the areas of human hand recognition and human-computer interaction. He has served as the Deputy Director of Shandong Provincial Key Laboratory of network based Intelligent Computing, group leader of Human Computer Interaction based on natural hand, editorial board member of Computer Aided Drafting Design and Manufacturing, CADDM, and also an editorial board member of The Open Virtual Reality Journal. He is a deputy editor of World Research Journal of Pattern Recognition and a member of Computer Graphics professional committee. Dr. Feng’s research interests are in human hand tracking/recognition/interaction, virtual reality, human-computer interaction, and image processing. His research has been extensively supported by the Key R\&D Projects of the Ministry of Science and Technology, Natural Science Foundation of China, Key Projects of Natural Science Foundation of Shandong Province, and Key R\&D Projects of Shandong Province with total grant funding over three million RMB. For more information, please refer to http://nbic.ujn.edu.cn/nbic/index.php.
\end{IEEEbiography}

\begin{IEEEbiography}[{\includegraphics[width=1in,height=1.25in,clip,keepaspectratio]{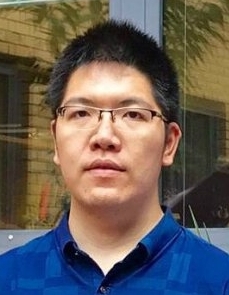}}]{Tianyang Xu} received the B.Sc. degree in electronic science and engineering from Nanjing University, Nanjing, China, in 2011. He received the PhD degree at the School of Artificial Intelligence and Computer Science, Jiangnan University, Wuxi, China, in 2019.
He is currently an Associate Professor at the School of Artificial Intelligence and Computer Science, Jiangnan University, Wuxi, China.
His research interests include visual tracking and deep learning.
He has published several scientific papers, including IJCV, ICCV, TIP, TIFS, TKDE, TMM, TCSVT etc. He achieved top 1 tracking performance in several competitions, including the VOT2018 public dataset (ECCV18), VOT2020 RGBT challenge (ECCV20), and Anti-UAV challenge (CVPR20).
\end{IEEEbiography}

\begin{IEEEbiography}[{\includegraphics[width=1in,height=1.25in,clip,keepaspectratio]{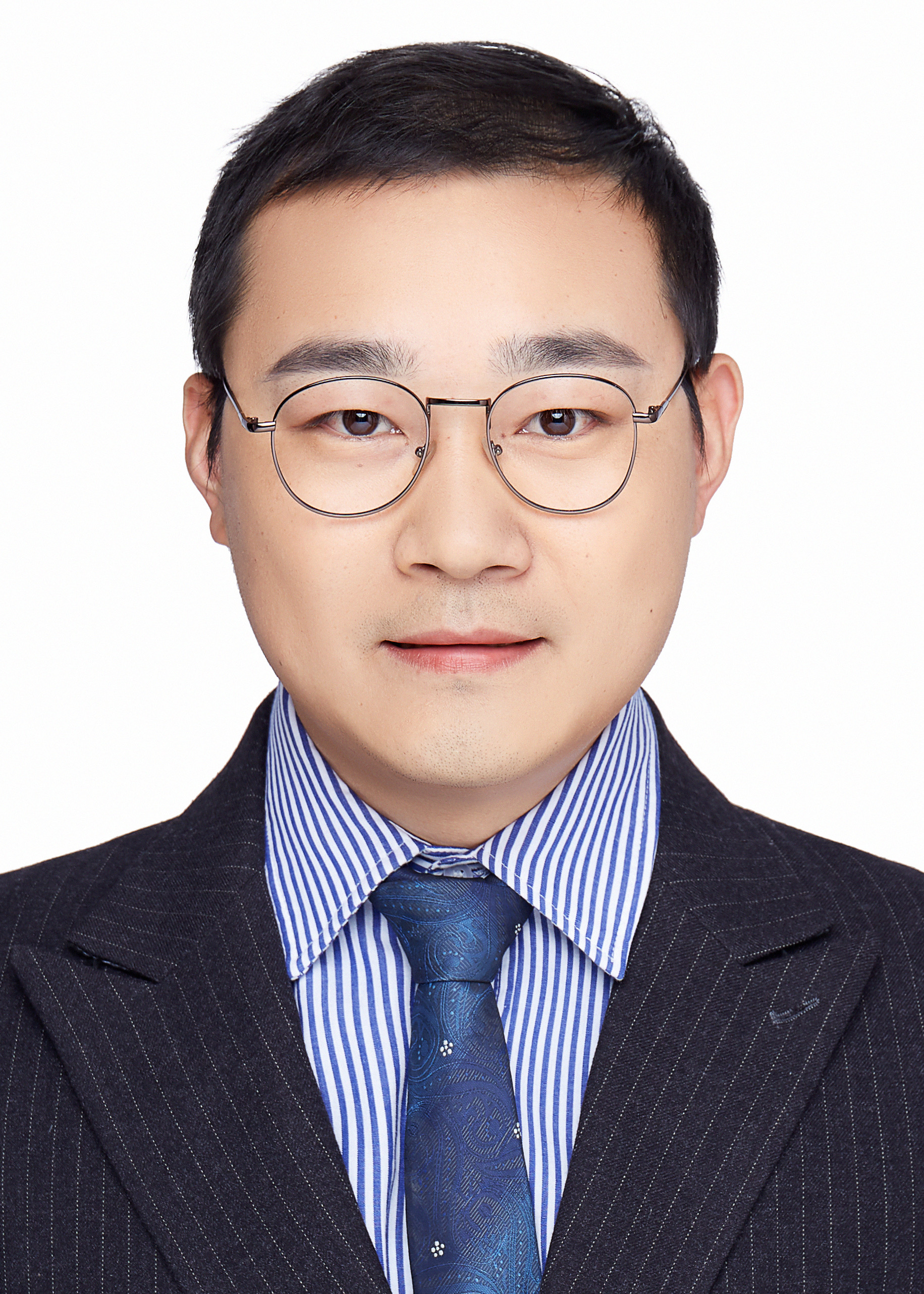}}]{Cong Hu} received the Ph.D. degree from Jiangnan University, Wuxi, China, in 2019. He is currently an Associate Professor with the School of Artificial Intelligence and Computer Science, Jiangnan University. He was a visiting Researcher from July 2018 to July 2019 with the Centre for Vision, Speech and Signal Processing, University of Surrey, Guildford GU2 7XH, UK. His research interests include pattern recognition, machine learning, and computer vision.
\end{IEEEbiography}

\begin{appendix}

\begin{figure*}[t]
\begin{minipage}{1\textwidth}
  \centering
  \includegraphics[trim=0mm 5mm 0mm 5mm, width=5.3in]{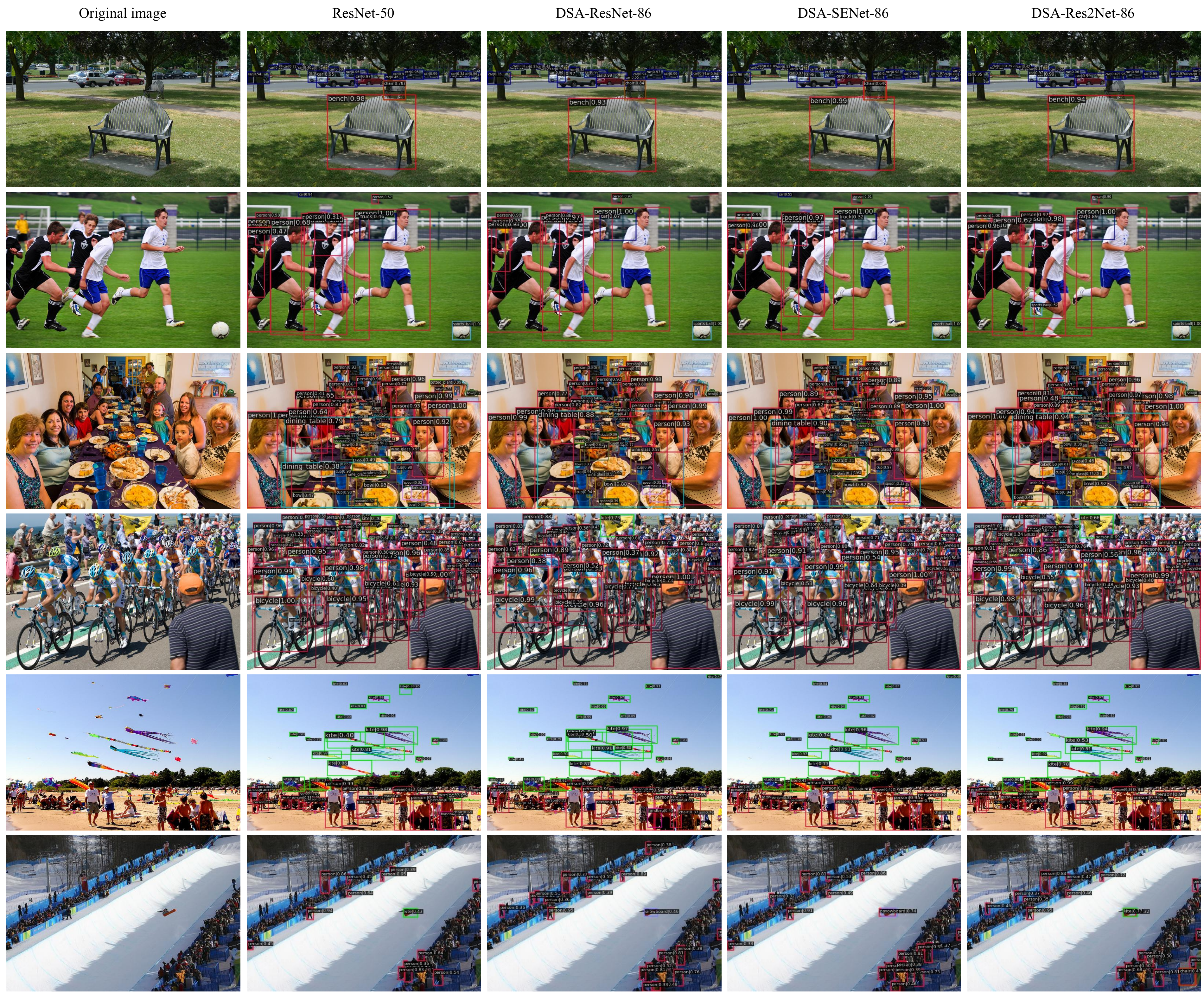}
  {\caption{Visualization of object detection results using ResNet-50, SDA-ResNet-86, SDA-SENet-86, and SDA-Res2Net-86 as backbone networks, respectively.}
  \label{fig:6}}
\end{minipage}
\end{figure*}

\begin{figure*}[t]
\begin{minipage}{1\textwidth}
  \centering
  \includegraphics[trim=0mm 5mm 0mm 5mm, width=5.3in]{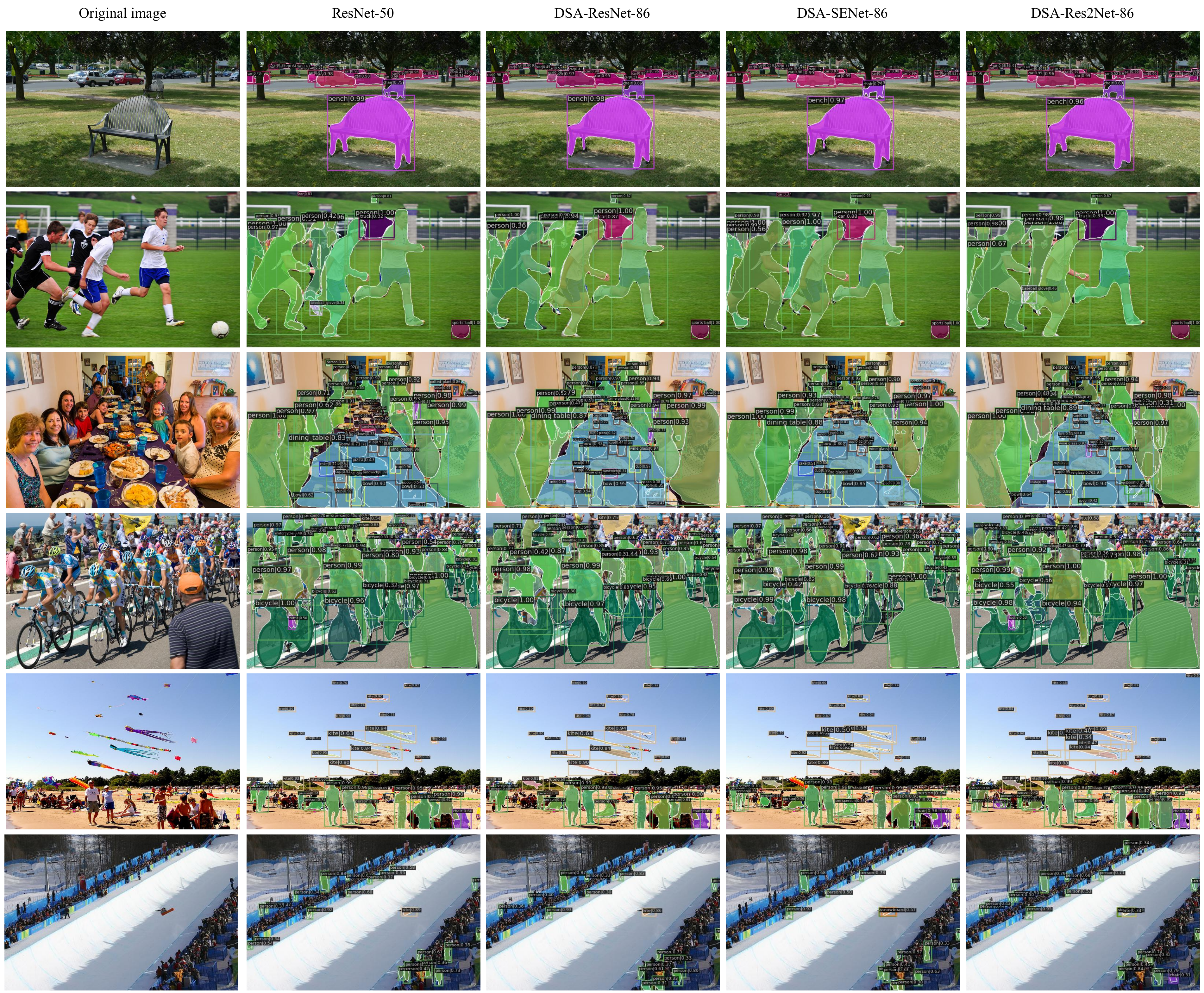}
  {\caption{Visualization of instance segmentation results using ResNet-50, SDA-ResNet-86, SDA-SENet-86, and SDA-Res2Net-86 as backbone networks, respectively.}
  \label{fig:7}}
\end{minipage}
\end{figure*}

\begin{figure*}[t]
\begin{minipage}{1\textwidth}
  \centering
  \subfigure[]{
  \includegraphics[trim=0mm 5mm 0mm 5mm, width=5.8in]{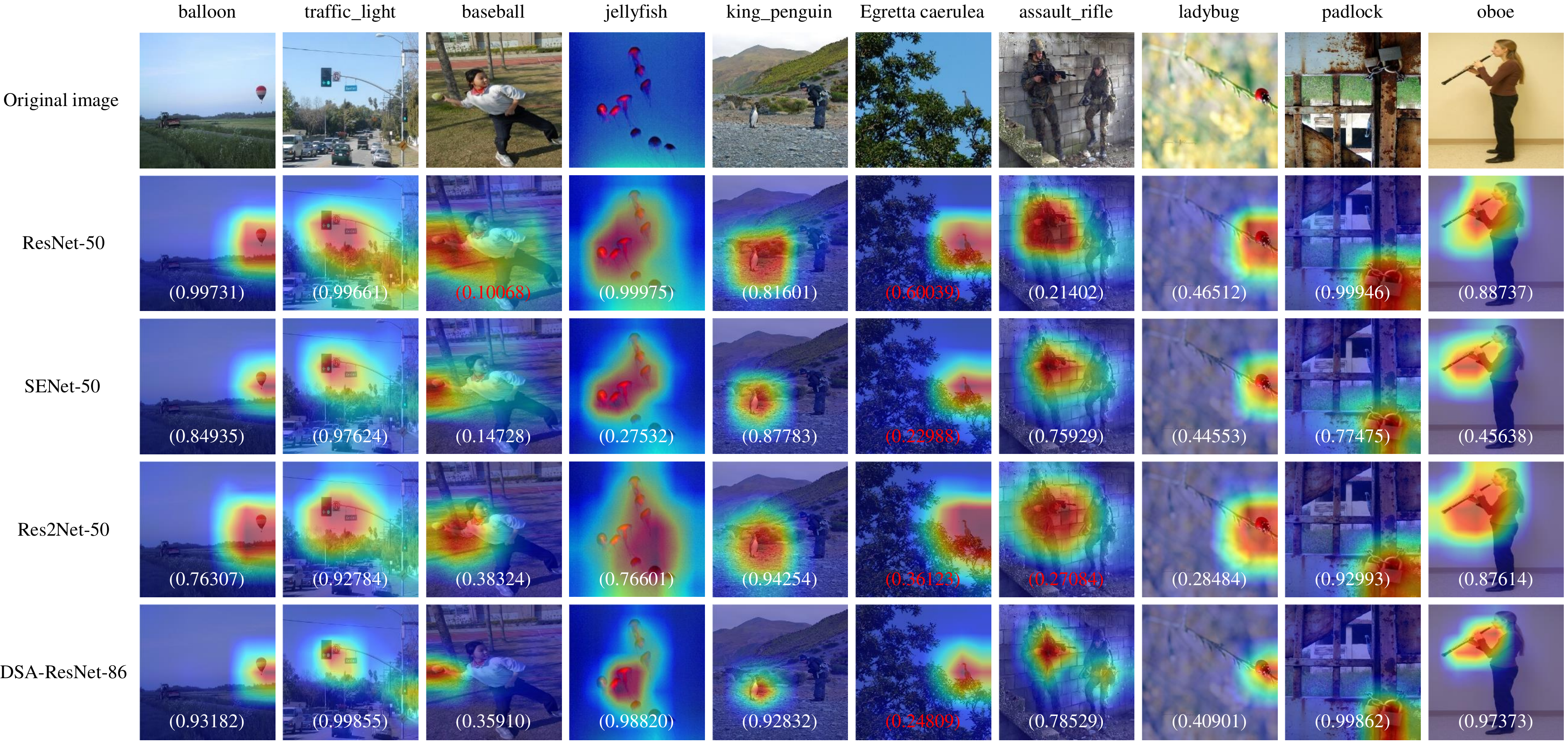}}

  \subfigure[]{
  \includegraphics[trim=0mm 5mm 0mm 5mm, width=5.8in]{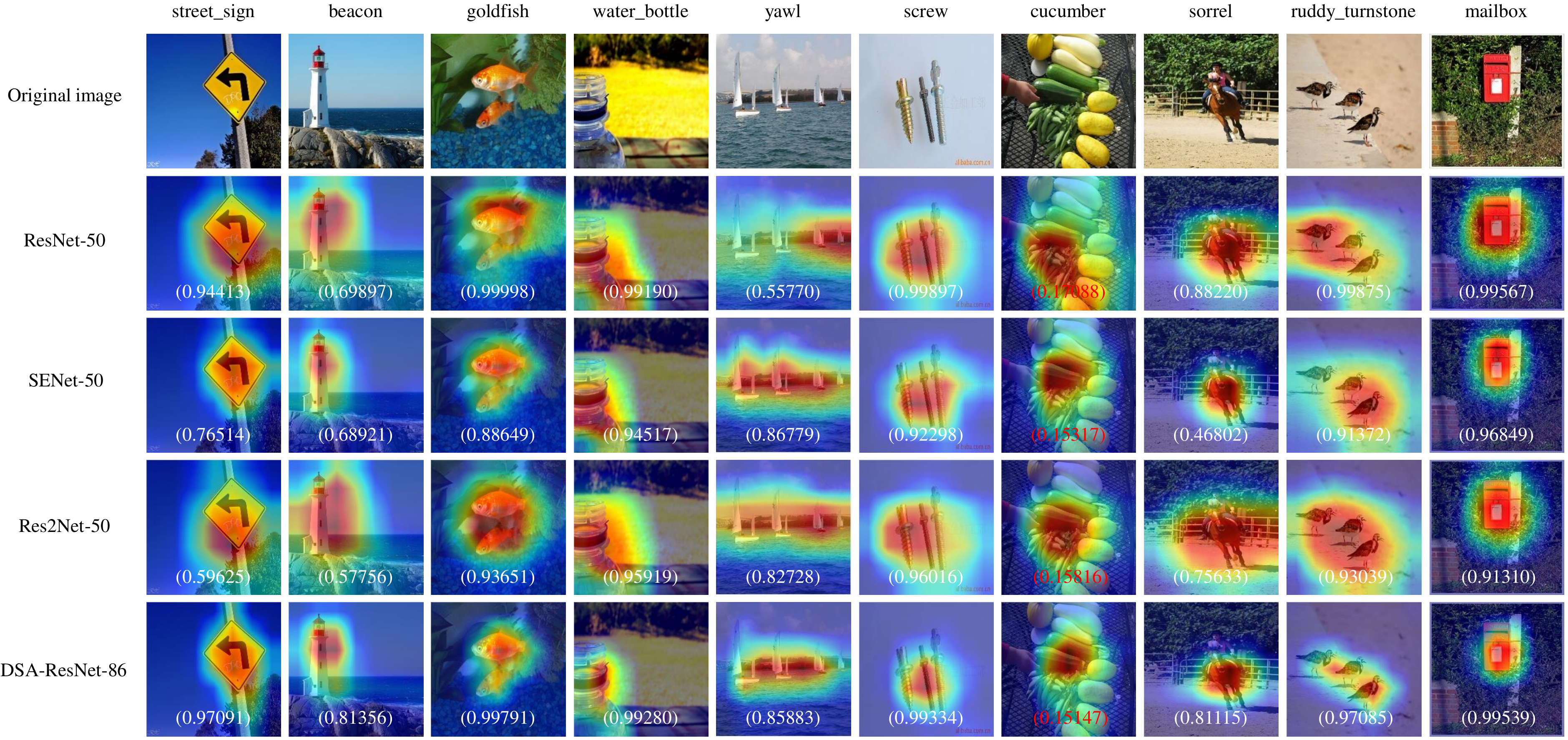}}

  \subfigure[]{
  \includegraphics[trim=0mm 5mm 0mm 5mm, width=5.8in]{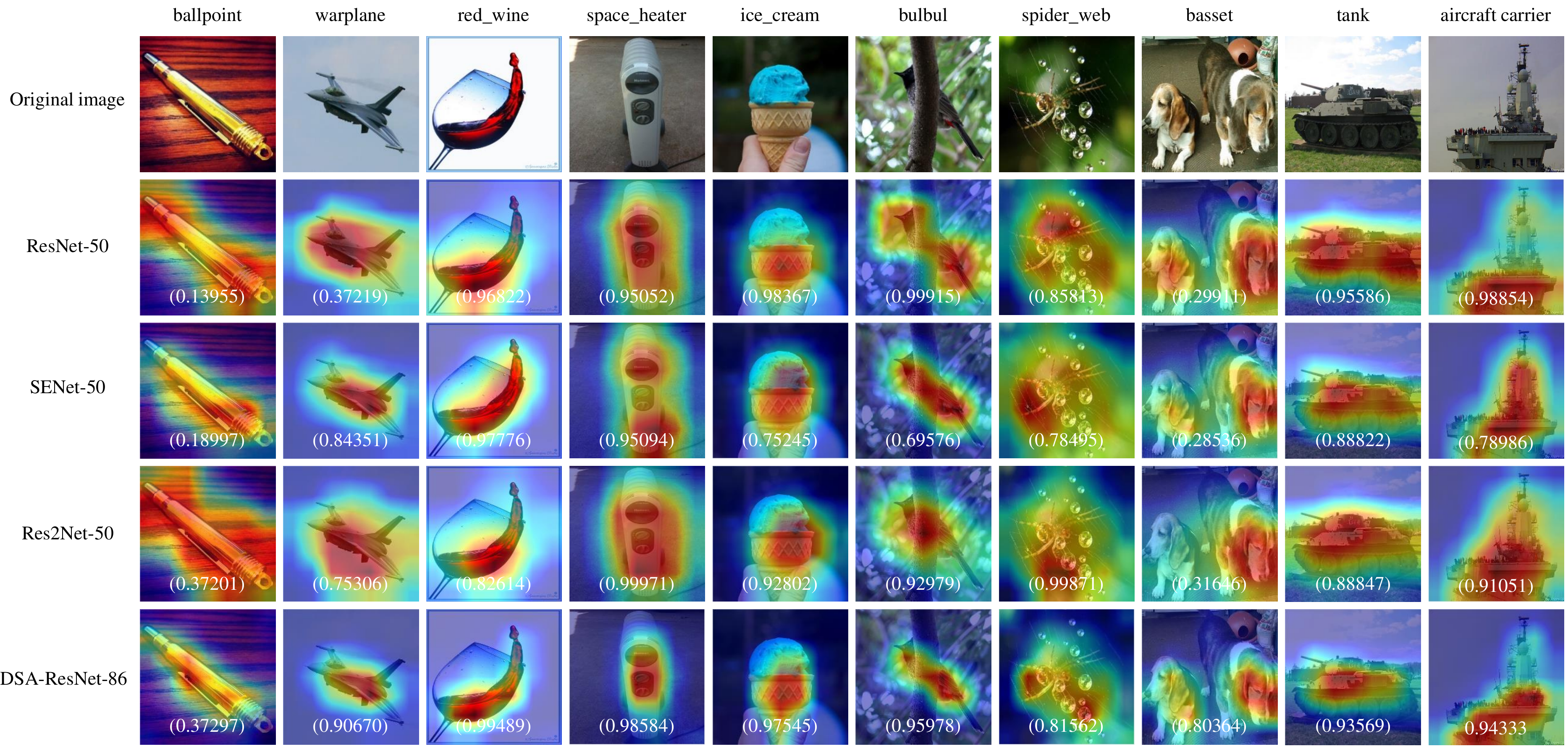}}
  {\caption{More Grad-CAM visualization results for different attention methods. (a) Small objects. (b) Medium objects. (c) Large objects. The softmax scores for the target class are displayed at the bottom of each image. The red scores denote the wrong classification results.}
  \label{fig:8}}
\end{minipage}
\end{figure*}

\end{appendix}

\end{document}